\documentclass[runningheads]{llncs}
\usepackage[T1]{fontenc}

\usepackage{graphicx}
\usepackage{microtype}
\usepackage{graphicx}
\usepackage{subfig}
\usepackage{booktabs} 
\usepackage[hyphens]{url}
\PassOptionsToPackage{hyphens}{url}\usepackage{hyperref}

\usepackage{amsmath}
\usepackage{amssymb}
\usepackage{mathtools}
\usepackage{float}
\usepackage{fancyhdr}
\usepackage{xcolor} 
\usepackage{algorithm}
\usepackage{algorithmic}
\usepackage[numbers]{natbib}
\usepackage{eso-pic} 
\usepackage{forloop}
\usepackage{url}

\usepackage{xcolor}         
\usepackage{color}

\usepackage{multirow}

\usepackage{amsmath,amsfonts,bm}

{}
{}
{}
{}

\def\eqref#1{equation~\ref{#1}}

\def\1{\bm{1}}

\def\ve{{\bm{e}}}

\def\vm{{\bm{m}}}

\def\vw{{\bm{w}}}
\def\vx{{\bm{x}}}
\def\vy{{\bm{y}}}

\def\mA{{\bm{A}}}

\def\mE{{\bm{E}}}

\def\mW{{\bm{W}}}

\DeclareMathAlphabet{\mathsfit}{\encodingdefault}{\sfdefault}{m}{sl}
\SetMathAlphabet{\mathsfit}{bold}{\encodingdefault}{\sfdefault}{bx}{n}

\newcommand{\Enc}{\mathtt{Enc}}
\newcommand{\Dec}{\mathtt{Dec}}

\newcommand{\VQ}{\mathtt{VQ}}
\newcommand{\sg}{\mathtt{sg}}

\newcommand{\hardmax}{\mathrm{hardmax}}

\DeclareMathOperator*{\argmin}{arg\,min}

\usepackage{graphicx}
\usepackage{rotating}

\newsavebox\CBox
\def\textBF#1{\sbox\CBox{#1}\resizebox{\wd\CBox}{\ht\CBox}{\textbf{#1}}}

\begin{document}
\sloppy

\title{Self-Organising Neural Discrete Representation Learning \`a la Kohonen}

\author{Kazuki Irie$^{\star}$\inst{1} \and
R\'obert Csord\'as$^{\star}$\inst{2} \and
J\"urgen Schmidhuber\inst{3,4}}
%

\authorrunning{K.~Irie, R.~Csord\'as, and J.~Schmidhuber}
%

\institute{Center for Brain Science, Harvard University, Cambridge, MA, USA \and
Stanford University, Stanford, CA, USA \and
The Swiss AI Lab, IDSIA, USI \& SUPSI, Lugano, Switzerland \and
AI Initiative, King Abdullah University of Science and Technology (KAUST), Thuwal, Saudi Arabia \\
\email{kirie@fas.harvard.edu,rcsordas@stanford.edu,juergen@idsia.ch}}
\maketitle              
\begin{abstract}

\renewcommand{\thefootnote}{\fnsymbol{footnote}} 
\footnotetext[1]{Equal contribution. Work done at IDSIA.}
\renewcommand{\thefootnote}{\arabic{footnote}} 
\setcounter{footnote}{0}

Unsupervised learning of discrete representations in neural networks (NNs) from continuous ones is essential for many modern applications.
Vector Quantisation (VQ) has become popular for this, in particular in the context of generative models, such as Variational Auto-Encoders (VAEs), where the exponential moving average-based VQ (EMA-VQ) algorithm is often used.
Here, we study an alternative VQ algorithm based on Kohonen's learning rule for the Self-Organising Map (KSOM; 1982).  EMA-VQ is a special case of KSOM. KSOM is known to offer two potential benefits:
empirically,
it
converges faster than EMA-VQ, and KSOM-generated
discrete representations form a topological structure on the grid whose nodes are the discrete symbols,
resulting in an artificial version of the brain's \textit{topographic map}.
We revisit these properties by using KSOM in VQ-VAEs for image processing.
In  our experiments, the speed-up compared to \textit{well-configured} EMA-VQ is only observable at the beginning of training, but KSOM is generally much more robust, e.g., w.r.t.~the choice of initialisation schemes.\footnote{Our code is public: \url{https://github.com/IDSIA/kohonen-vae}.\looseness=-1}\footnote{Accepted to ICANN 2024. The Version of Record of this contribution is published by Springer. An earlier version was presented at ICML 2023 Workshop on Sampling and Optimization in Discrete Space (SODS).}

\keywords{self-organizing maps \and Kohonen maps \and vector quantisation \and VQ-VAE \and discrete representation learning}
\end{abstract}

\section{Introduction}
\label{sec:intro}
Internal representations in artificial neural networks (NNs) are
continuous-valued vectors.
As such, they are rarely identical in different contexts.
In many scenarios, however, it is natural and desirable to treat some of these non-identical but similar vectors as representing a common discrete symbol from a fixed-size codebook/lexicon shared across various contexts \citep{OordVK17,HuMTMS17}.
Such \textit{discrete representations} would allow us to express and manipulate hidden representations of NNs using a set of symbols.
Unsupervised learning of such discrete representations is motivated in various contexts.
For example, in certain algorithmic or reasoning tasks (e.g., \cite{liska2018memorize, hupkes2018learning, ctl2022}), learning of such representations may be a key for generalisation, since intermediate results in such tasks are inherently discrete.

Recently, discrete representation learning
via \textit{Vector Quantisation} (VQ) has become
popular for practical reasons too.
The Vector Quantised-Variational Auto-Encoders (VQ-VAEs) by van den Oord et al.~\cite{OordVK17} use
VQ as a pre-processing step to
\textit{tokenise} high-dimensional data, such as images,
i.e., to represent an image as a sequence of discrete symbols.
The resulting sequence can then be processed by a powerful sequence processor, e.g., a Transformer variant \citep{trafo, Schmidhuber:91fastweights, schlag2021linear}.
Similar techniques have been applied to other kinds of data such as video \citep{walker2021predicting, yan2021videogpt} and audio \citep{BaevskiSA20, dhariwal2020jukebox, TjandraS020, borsos2022audiolm}).
Today, many large-scale text-to-image systems such as \textit{DALL-E} \citep{RameshPGGVRCS21},  \textit{Parti} \citep{yu2022scaling}, or \textit{Latent Diffusion Models} \citep{RombachBLEO22}, also use VQ in one way or another.\looseness=-1

Here we study the learning rules of Kohonen Self-Organising Maps or Kohonen Maps \citep{kohonen1982self, kohonen2001book} as the VQ algorithm for discrete representation learning in NNs.
For shorthand, we refer to the corresponding algorithm as \textit{KSOM} (reviewed in Sec.~\ref{sec:background}).
In fact, KSOM is a classic VQ algorithm \citep{nasrabadi1988vector}, and the exponential moving average-based VQ (EMA-VQ; \cite{OordVK17, RazaviOV19, KaiserBRVPUS18, roy2018towards}) commonly used today is a special case thereof (Sec.~\ref{sec:method}).
KSOM is known to offer two potential benefits over EMA-VQ.
First, KSOM is empirically reported to perform faster VQ (see, e.g., \cite{de2004use}).
Second, discrete representations learned by KSOM form a \textit{topological} structure in the pre-specified \textit{grid} whose nodes represent the discrete symbols from the codebook.
Such a grid is typically one- or two-dimensional, and symbols that are spatially close to each other on the grid represent features that are close to each other in the original input space.
This \textit{topological mapping} property has helped practitioners in certain applications to visualise/interpret their data (see, e.g.~\cite{tirunagari2016visualisation}).
While such a property is arguably of limited importance in today's deep learning,
Kohonen's algorithm is specifically designed to achieve this property which is known in the brain as \textit{topographic organisation}.
KSOM allows to naturally achieve an artificial version thereof, as a by-product of the VQ algorithm.

We explore these properties in modern NNs by using KSOM as the VQ algorithm in VQ-VAEs \citep{OordVK17, RazaviOV19} for image processing.
Importantly, we also revisit the configuration details of the baseline EMA-VQ (e.g., initialisation of EMAs).
We show that proper configurations are crucial for EMA-VQ to optimally perform, while KSOM is robust, and performs well in all cases.

\section{Background: Kohonen Maps}
\label{sec:background}
Here we provide a brief review of Kohonen's Self-Organising Maps (KSOMs).

\subsection{(Online) Algorithm}
\label{subsec:online_algo}
Teuvo Kohonen (1934-2021)'s Self-Organising Map \citep{kohonen1982self} is an unsupervised learning/clustering algorithm which achieves both \textit{vector quantisation} (VQ) and \textit{topological mapping} (Sec.~\ref{subsec:topo}).
Let $T$, $K$, $d_{\text{in}}$ and $t$ denote positive integers.
The algorithm requires to define a \textit{distance function} $\delta: \mathbb{R}^{d_{\text{in}}} \times \mathbb{R}^{d_{\text{in}}} \rightarrow \mathbb{R}_{\geq 0}$ as well as a \textit{neighbourhood matrix} $\mA \in \mathbb{R}^{K \times K}$ with $0 \leq \mA_{i, j} \leq 1$ for all $i,j \in \{1,..., K\}$, whose roles are specified later.
We consider $T$ input vectors\footnote{\normalsize Here we use $T$ as both the number of inputs and iterations.}
$(\vx_1, ..., \vx_T)$ with $\vx_t \in \mathbb{R}^{d_{\text{in}}}$ for $t \in \{1,..., T\}$,
and a weight matrix $\mW \in \mathbb{R}^{K \times d_\text{in}}$ which we describe as a list of $K$ weight vectors $(\vw_{1},..., \vw_{K}) = \mW^{\intercal}$ with $\vw_{k} \in \mathbb{R}^{d_{\text{in}}}$ for $k \in \{1,..., K\}$ representing a \textit{codebook} of size $K$.
The algorithm clusters input vectors into $K$ clusters where the prototype (or the centroid) of cluster $k \in \{1,..., K\}$ is $\vw_{k}$.
At the beginning, these weight vectors are randomly initialised as $(\vw_{1}^{(0)},..., \vw_{K}^{(0)})$ where the super-script denotes the iteration step.
The KSOM algorithm learns these weight vectors iteratively as follows.

For each step $t \in \{1,..., T\}$, we process an input $\vx_t$ by computing the index $k^*$ (typically called \textit{best matching unit}) of the weight vector that is the closest to the input $\vx_t$ according to $\delta$, i.e.,
\begin{align}
\label{eq:find_max}
k^* &= \underset{1 \leq k \leq K}{\argmin} \, \delta(\vx_t, \vw_k^{(t-1)})
\end{align}
then the weights are updated; for all $k \in \{1,..., K\}$,
\begin{align}
\label{eq:update_weights}
\vw_k^{(t)} &= \vw_k^{(t-1)} + \beta \mA_{k^*, k}^{(t-1)} (\vx_t - \vw_k^{(t-1)})
\end{align}
where $\beta \in \mathbb{R}_{>0}$ is the learning rate, and the super-script added to $\mA_{k^*, k}^{(t)}$ indicates that it also changes over time.

Now, we need to specify the distance function $\delta$ in Eq.~\ref{eq:find_max} and the neighbourhood matrix $\mA^{(t)}$ in Eq.~\ref{eq:update_weights}.

\textbf{Distance function $\delta$.} A typical choice for $\delta$ is the Euclidean distance which we also use in all our experiments.
Strictly speaking, $\delta$ does not have to be a metric; any kind of (dis)similarity function can be used, e.g., negative dot product is also a common choice (\textit{dot-SOM} \citep{kohonen2001book}).

\textbf{Grid of Codebook Indices.}
In order to define the neighbourhood matrix $\mA^{(t)}$, we first need
to define a \textit{grid} or lattice of the codebook indices on which \textit{neighbourhoods} are defined.
Let $D$ denote a positive integer.
We define a map $\mathcal{G}: \{1,..., K\} \rightarrow \mathbb{N}^{D} $ which maps each codebook index $k \in \{1,..., K\}$ to its Cartesian coordinates $\mathcal{G}(k) \in \mathbb{N}^{D}$ in the $D$-dimensional space representing the grid in question.
Typically, the grid is 1D or 2D ($D=$ $1$ or $2$).
In the 1D case, the original index $k \in \{1,..., K\}$ and its coordinates on the map $\mathcal{G}(k)$ are the same: $\mathcal{G}(k)=k$.
In the 2D case, $\mathcal{G}(k)$ corresponds to the Cartesian $(x, y)$-coordinates on a 2D-rectangular grid formed by $K$ nodes (assuming that $K$ is chosen such that this is possible).
Once $\mathcal{G}$ is defined, one can measure the ``distance'' between two codebook indices $i, j \in \{1,..., K\}$ as their Euclidean distance on the grid, i.e., $\|\mathcal{G}(i) - \mathcal{G}(j)\|$.
This finally allows us to define the neighbourhood:
given a pre-defined threshold distance $d(\mathcal{G}) \in \mathbb{R}_{\geq 0}$,
for $i,j \in \{1,..., K\}$, $i$ is within the neighbourhood of $j$ on the map $\mathcal{G}$ if and only if $\|\mathcal{G}(i) - \mathcal{G}(j)\| \leq d(\mathcal{G})$.

\begin{figure}[h]
\centering

\subfloat[Illustration of a hard neighborhood.]{
            \centering
		\hspace{5mm}\includegraphics[width=.145\linewidth]{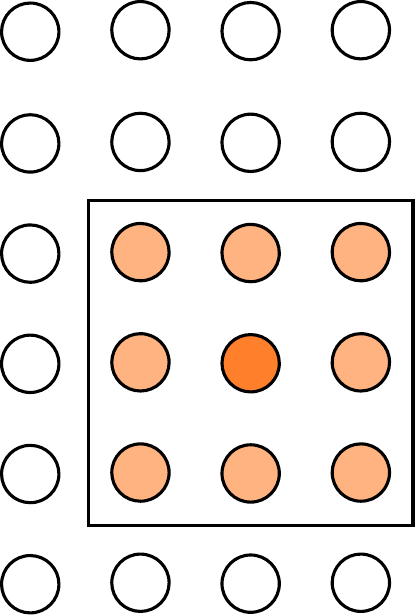}\hspace{5mm}
            \label{fig:hard_neighborhood}
	}
 \hfill
	\subfloat[Gaussian neighborhood: Early training.]{
            \centering
		\includegraphics[width=.25\linewidth]{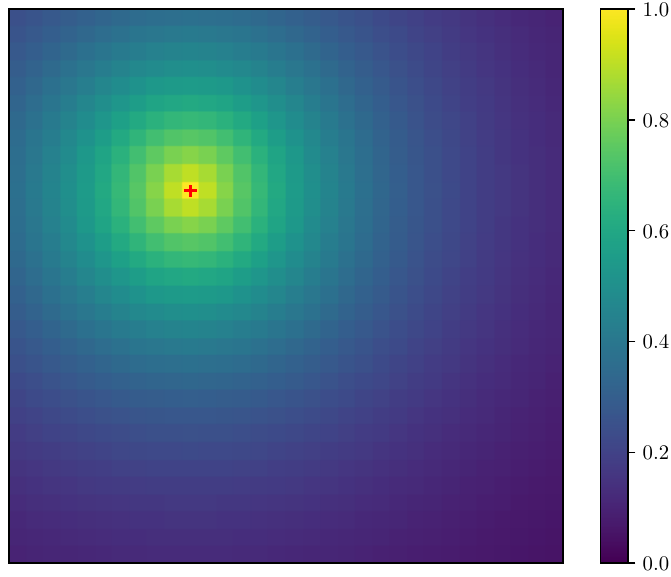}
	}
 \hfill
	\subfloat[Gaussian neighborhood: Mid training.]{
            \centering
		\includegraphics[width=.25\linewidth]{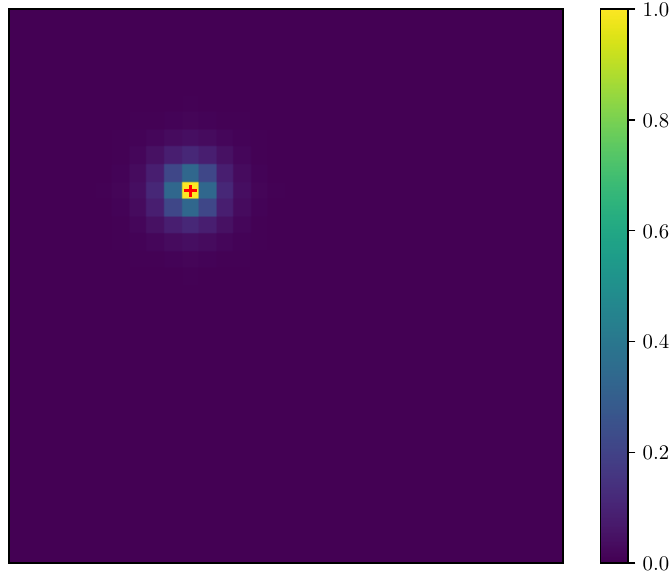}
	}
	\caption{(a) Illustration of \textit{hard} neighbourhoods (Eq.~\ref{eq:hard}) in the 2D case. This is a 6x4 grid with $K=24$ nodes. Considering the left-bottom corner node as the origin (0, 0), the eight neighbours of the node (2, 2) are highlighted. (b) and (c): Illustration of \textit{Gaussian} neighbourhoods in the 2D case with shrinking (Eq.~\ref{eq:gauss}) at two different stages of training.}
	\label{fig:gaussian_neighborhood}
\end{figure}

\textbf{Neighbourhood Matrix $\mA^{(t)}$.}
In Eq.~\ref{eq:update_weights}, the coefficient of the neighbourhood matrix $\mA_{k^*, k}^{(t)}$ has the role of adapting the weights/strengths of updates for each cluster $k$ according to its distance from the best matching unit $k^*$ on the map $\mathcal{G}$: as can be seen in Eq.~\ref{eq:update_weights}, $\beta \mA_{k^*, k}^{(t)}$ is the effective learning rate for the update.
Essentially, the best matching unit obtains the full update $\mA_{k^*, k^*}^{(t)} = 1$, while the updates for all others ($k \neq k^*$) are scaled down by $0 \leq \mA_{k^*, k}^{(t)} \leq 1$.
In practice, there are various ways to define such an $\mA^{(t)}$.
Here we focus on two variants: \textit{hard} and \textit{Gaussian} neighbourhoods.

Let us first define the \textit{hard} variant \textit{without} dependency on the time index $t$.
In the \textit{hard} variant, for all $i \in \{1,..., K\}$, $\mA_{i, i} = 1$, and for all $i,j$ such that $i \neq j$,\looseness=-1
\begin{align}
\label{eq:hard}
\mA_{i, j}=
\begin{cases}
1 & \text{if} \,\, \|\mathcal{G}(i) - \mathcal{G}(j)\| \leq d(\mathcal{G}) \\
0 & \text{otherwise}
\end{cases}
\end{align}
In the 1D case, setting $d(\mathcal{G})=1$ yields two neighbour indices.
Similarly, in the 2D case, $d(\mathcal{G})=\sqrt{2}$ defines the eight surrounding nodes as the neighbours,
which are illustrated in Figure \ref{fig:hard_neighborhood}.

In practice, we introduce an extra dependency on time $t$
with the goal of \textit{shrinking} the neighbourhood over time.
For all $i \in \{1,..., K\}$, $\mA_{i, i}^{(t)} = 1$, and for all $i,j$ such that $i \neq j$,
\begin{align}
\label{eq:shrink}
\mA_{i, j}^{(t)} =
\begin{cases}
1 / (1 + t * \tau)& \text{if} \,\, \|\mathcal{G}(i) - \mathcal{G}(j)\| \leq d(\mathcal{G}) \\
0 & \text{otherwise}
\end{cases}
\end{align}
where $\tau \in \mathbb{R}_{>0}$ is a hyper-parameter representing the \textit{shrinking step} which controls the speed of shrinking: larger $\tau$ implies faster shrinking.
Shrinking in KSOM is important to obtain good performance at convergence \citep{kohonen2001book}.

In the alternative, \textit{Gaussian} variant, $\mA_{i, j}^{(t)}$ is expressed as an exponential function of the distance between indices on $\mathcal{G}$:
\begin{align}
\label{eq:gauss}
\mA_{i, j}^{(t)} =
\exp(-\|\mathcal{G}(i) - \mathcal{G}(j)\|^2 / \sigma(t)^2)
\end{align}
where for $\sigma(t) \in \mathbb{R}$ we take $\sigma(t)^2 = 1 / (1 + t * \tau)$.
Note that, also in this case, $\mA^{(t)}$ reduces to an identity matrix when $t \rightarrow +\infty$.
Figure \ref{fig:gaussian_neighborhood} provides an illustration.\looseness=-1

\textbf{Relation to Hebbian Learning.}
We note that the online algorithm of Eqs.~\ref{eq:find_max}-\ref{eq:update_weights} can be also 
interpreted (see also \cite{yin2008self} on this relation) as a variant of Hebbian learning \citep{hebb1949organization}.
Using the dot product-based similarity function for $\delta$ in Eq.~\ref{eq:find_max}, and by defining $\hardmax: \mathbb{R}^{K} \rightarrow \mathbb{R}^{K}$ as the function that outputs 1 for the largest entry of the input vector, and 0 for all others, 
we can express Eqs.~\ref{eq:find_max}-\ref{eq:update_weights} in a fully matrix-form by defining a one-hot vector $\vy_t \in \mathbb{R}^{K}$,
\begin{align}
\label{eq:find_max_mat}
\vy_t &= \hardmax(\mW_{t-1} \vx_t) \\
\label{eq:update_weights_mat}
\mW_t &= \mW_{t-1} + (\beta \mA^{(t-1)} \vy_t) \otimes (\vx_t - \mW_{t-1}^{\intercal}\vy_t)
\end{align}
where $\otimes$ denotes outer product.
With the neighbourhood reduced to zero, this is essentially the \textit{winner-take-all} Hebbian learning.
While this relation is not central to this work, we come back to this when we discuss the differentiable version of KSOM later in Sec.~\ref{sec:discussion}.

In passing, we also note that the last term in Eqs.~\ref{eq:update_weights} and \ref{eq:update_weights_mat} corresponds to Oja's \cite{oja1982simplified}
\textit{forgetting term} which is not part of the original 1982 algorithm \citep{kohonen1982self} but has been added later (see, e.g., \cite{kohonen2001book}).

\subsection{Batch Algorithm \& Relation to K-means}
\label{sec:batch_algo}

The algorithm described above is an \textit{online} algorithm which updates weights after every input.
The \textit{batch} version thereof (\cite{kohonen1999comparison, kohonen2001book}; see also \cite{cottrell2018self}), which takes into account all data points $(\vx_1, ..., \vx_T)$ for a single update of $\{\vw_{1},..., \vw_{K}\}$, can be defined as follows.

At each iteration step $t$, we compute the best matching unit for each input $x_i$ for all $i \in \{1,.., T\}$ (we now use the sub-script $i$ to index data points not to confuse it with the iteration index $t$).
Each input $x_i$ is a \textit{member} of one of the $K$ clusters.
The results can be summarised for each cluster $k \in \{1,.., K\}$ as the set $\mathcal{C}_k^{(t)}$ containing indices of its members at step $t$. We denote its cardinality 
by $|\mathcal{C}_k^{(t)}|$.
The batch algorithm updates the weight vector $\vw_k^{(t)}$ for cluster $k$ as the average of its members and their neighbours weighted by the neighbourhood coefficients.
That is, $\vw_k^{(t)}$ is computed as the quotient of the sum $\vm_k^{(t)}$ of all inputs belonging to the corresponding cluster $k$ and their neighbours weighted by the neighbourhood coefficients, and the corresponding weighted count  $N_k^{(t)}$, i.e.,\\
\begin{minipage}{0.49\linewidth}
    \begin{equation}
     \vm_k^{(t)} = \sum_{j=1}^K \mA_{j, k}^{(t-1)} \sum_{i \in \mathcal{C}_j^{(t)}}\vx_i 
\label{eq:numerator}
    \end{equation}
\end{minipage}%
\begin{minipage}{0.51\linewidth}
    \begin{equation}
\label{eq:denominator}
N_k^{(t)} = \sum_{j=1}^K \mA_{j, k}^{(t-1)} |\mathcal{C}_j^{(t)}|
    \end{equation}
\end{minipage}%
    \begin{equation}
\label{eq:update_final}
\vw_k^{(t)} =  \vm_k^{(t)} / N_k^{(t)}
    \end{equation}

\begin{remark}[Relation to K-means]
\label{remark:k-means}
In the case where the \textit{neighbourhood is reduced to zero}, i.e., for all $t$, 
$\mA_{i, i}^{(t)} = 1$ and for all $i,j$ such that $i \neq j$, $\mA_{i, j}^{(t)} = 0$,
Eqs.~\ref{eq:numerator}-\ref{eq:update_final} reduce to the standard \textit{K-means} algorithm \citep{lloyd1982least}.
Similarly, in such a case, Eqs.~\ref{eq:find_max}-\ref{eq:update_weights} reduce to an \textit{online K-means} algorithm \citep{macqueen1967classification}.
\end{remark}

For deep learning applications, the algorithm needs to be both \textit{online} and \textit{mini-batch};
we discuss the corresponding extensions in Sec.~\ref{sec:method}.

\subsection{Topographical Maps in the Brain as Motivation}
\label{subsec:topo}
Above we describe how KSOM performs clustering, i.e., \textit{vector quantisation}.
Here we discuss another property of this algorithm which is \textit{topological mapping}.

It is known that there are multiple levels of \textit{topographical maps} in the brain, e.g.,
different regions of the brain specialise to different types of sensory inputs (vision, audio, touch, etc), and e.g., within the somatosensory part, regions that are responsible for different parts of the body are \textit{ordered} according to the anatomical order in the body.
A famous illustration of this is the 
``sensory homunculus.''\looseness=-1

The design of KSOM is inspired by such topographical maps.
Many have proposed computational mechanisms to achieve such a property in the 1970s/80s \citep{von1973self, von1977label, willshaw1979marker, willshaw1976patterned, amari1980topographic}.
Kohonen \cite{kohonen1982self} achieves this by introducing the concept of neighbourhoods between the (output) neurons.
In the algorithm above (Sec.~\ref{subsec:online_algo}), all output neurons first compete against each other (Eq.~\ref{eq:find_max}) to yield a winner neuron. Then, the update is distributed to neurons that are spatially close to the winner through the coefficients of the neighbourhood matrix (Eqs.~\ref{eq:update_weights} and \ref{eq:numerator}).
As a result, clusters whose indices are spatially close on the grid are encouraged to store inputs that are close to each other in the feature space.

This is an unconventional feature for artificial NNs, since unlike the biological ones,
artificial NNs do not have any physical constraints; there is no geometry nor distance between neurons. 
KSOM's neighbourhoods introduce such a structure.
From the machine learning perspective, the resulting topological ordering has limited practical benefits.
Even if it may potentially facilitate interpretation via direct visualisation, other embedding visualisation tools could fit the bill equally well.
From the neuroscience perspective, however, it may be a property that contributes in filling the gap between artificial NNs and the biological ones (see also \cite{constantinescu2016organizing}).\looseness=-1

\section{Alternative VQ in VQ-VAEs}
\label{sec:method}

The general idea of this work is to replace the VQ algorithm used in standard VQ-VAEs by Kohonen's algorithm (Sec.~\ref{sec:background}).
While the method can be applied to various data modalities, we focus on image processing as a representative example.\looseness=-1

\subsection{Background: VQ-VAEs}
\label{sec:vq_vae}
Let $d_{\text{in}}, d_{\text{emb}}, N, K$ be positive integers.
A VQ-VAE \citep{OordVK17} consists of an encoder NN, $\Enc: \mathbb{R}^{d_{\text{in}}} \rightarrow \mathbb{R}^{N \times d_{\text{emb}}}$, a decoder NN, $\Dec: \mathbb{R}^{N \times d_{\text{emb}}} \rightarrow \mathbb{R}^{d_{\text{in}}}$, and a codebook of size $K$ whose weights are $(\vw_{1},..., \vw_{K})$ with $\vw_{k} \in \mathbb{R}^{d_{\text{emb}}}$ for $k \in \{1,..., K\}$.
The encoder transforms an input $\vx \in \mathbb{R}^{d_{\text{in}}}$ to a sequence of $N$ embedding vectors $(\ve_1,..., \ve_N) = \mE$ with $\ve_i \in \mathbb{R}^{d_{\text{emb}}}$ for $i \in \{1,..., N\}$.
Each of these embeddings is quantised to yield $(\VQ(\ve_1),..., \VQ(\ve_N)) = \mE'$ with $\VQ(\ve_i) \in \mathbb{R}^{d_{\text{emb}}}$ for $i \in \{1,..., N\}$ where $\VQ$ denotes the VQ operation.
The decoder transforms the quantised embeddings $\mE'$ to a reconstruction of the original input $\hat{\vx} \in \mathbb{R}^{d_{\text{in}}}$.
The corresponding operations can be expressed as follows.
\begin{align}
(\ve_1,..., \ve_N) &= \Enc(\vx) \\
\text{For all} \,\, i \in \{1,..., N\},\,\, k_i^* &= \underset{1 \leq k \leq K}{\argmin} \, \|\ve_i - \vw_k \|_2 \\
\mE' = (\VQ(\ve_1),..., \VQ(\ve_N)) &= (\vw_{k_1^*}, ..., \vw_{k_N^*}) \\
\hat{\vx} &= \Dec(\mE')
\end{align}
The parameters of the encoder and decoder are trained to minimise:
\begin{align}
\label{eq:loss}
\|\vx - \hat{\vx}\|_2^2 + \lambda \dfrac{1}{N}\sum_{i=1}^N\|\sg(\vw_{k_i^*}) - \ve_i \|_2^2
\end{align}
where the first term is the reconstruction loss, and the second term is the so-called \textit{commitment loss}, weighted by a hyper-parameter $\lambda \in \mathbb{R}_{>0}$, which encourages the encoder to output embeddings that are close to their quantised counterparts ($\sg$ denotes ``stop gradient'' operation to prevent gradients to propagate through $\vw_{k_i^*}$).
As noted in \cite{OordVK17},
by assuming a uniform prior over the discrete latents, the standard KL term of the VAE loss \citep{KingmaW13} can be omitted as a constant.
In the reconstruction term, as the quantisation operation is non-differentiable,
the straight-through estimator \citep{hinton2012neural, bengio2013estimating} is used, i.e., gradients are directly copied from the decoder input to the encoder output.

The weights of the codebook prototypes $(\vw_{1},..., \vw_{K})$ are trained by a variant of online mini-batch K-means algorithm which keeps track of exponential moving averages (EMAs) of two quantities for each cluster $k$: the sum of updates $\vm_k^{(t)}$, and the count of members in the cluster $N_k^{(t)}$.
Their quotient yields the estimate of the weights at step $t$. Using the same notation $\mathcal{C}_k^{(t)}$ as in Sec.~\ref{sec:batch_algo} for the set of encoder output indices (ranging from 1 to $N*B$ where $B$ is a positive integer denoting the batch size) that are members of cluster $k$ at step $t$, and by denoting encoder outputs in the batch as $(\ve_1,..., \ve_{N*B})$, it yields:
\begin{align}
\label{eq:vq_vae_sum}
\vm_k^{(t)} &= (1-\beta) \vm_k^{(t-1)} + \beta \sum_{i \in \mathcal{C}_k^{(t)}}\ve_i^{(t)} \\
\label{eq:denominator_vq_vae}
N_k^{(t)} &= (1-\beta) N_k^{(t)} + \beta |\mathcal{C}_k^{(t)}| \\
\label{eq:vq_vae_update}
\vw_k^{(t)} &=  \vm_k^{(t)} / N_k^{(t)}
\end{align}
where $(1-\beta)$ is the EMA decay typically set to $0.99$ (i.e., $\beta = 0.01$).
For shorthand, we refer to this algorithm as EMA-VQ.

While it is also possible to train these codebook parameters
through gradient descent by introducing an extra term in the loss of Eq.~\ref{eq:loss} (batching is omitted for consistency): $1 / N \sum_{i=1}^N\|\vw_{k_i^*} - \sg(\ve_i) \|_2^2$, van den Oord et al.~\cite{OordVK17} recommend the EMA-based approach above, and many later works \citep{RazaviOV19, OzairLRAOV21} follow this practice, while in VQ-GANs \citep{EsserRO21, YuLKZPQKXBW22}, the gradient-based approach is also commonly used.

\subsection{Kohonen-VAEs}
\label{sec:k-vae}
We use KSOM (Sec.~\ref{sec:background}) as the VQ algorithm to learn the codebook weights in VQ-VAEs (Sec.~\ref{sec:vq_vae}).
Essentially, we replace the EMA-VQ algorithm of Eqs.~\ref{eq:vq_vae_sum}-\ref{eq:vq_vae_update} by an \textit{online mini-batch} version of KSOM.
Such an algorithm can be obtained by introducing exponential moving averaging into the batch version of KSOM (Eqs.~\ref{eq:numerator}-\ref{eq:update_final}) with a decay of $1 - \beta$.
That is, we keep track of EMAs of both the weighted sum of the updates ($\vm_k^{(t)}$; Eq.~\ref{eq:numerator}) and the weighted count of members ($N_k^{(t)}$; Eq.~\ref{eq:denominator}) for each cluster $k \in \{1,..., K\}$ (where weights are the neighbourhood coefficients).
Their quotient yields the estimate of the weights of codebook prototypes at step $t$.
Using the same notations as in Sec.~\ref{sec:vq_vae}, i.e., 
$\mathcal{C}_k^{(t)}$ denotes the set of indices of encoder outputs that are members of cluster $k$ at step $t$, and $(\ve_1,..., \ve_{N*B})$ denotes the encoder outputs in the batch, it yields:
\begin{align}
\label{eq:numerator_kvae}
\vm_k^{(t)} &= (1-\beta) \vm_k^{(t-1)} + \beta \sum_{j=1}^K \mA_{j, k}^{(t-1)} \sum_{i \in \mathcal{C}_j^{(t)}}\ve_i^{(t)} \\
\label{eq:denominator_kvae}
N_k^{(t)} &= (1-\beta) N_k^{(t)} + \beta \sum_{j=1}^K \mA_{j, k}^{(t-1)} |\mathcal{C}_j^{(t)}| \\
\label{eq:quotient_kvae}
\vw_k^{(t)} &=  \vm_k^{(t)} / N_k^{(t)}
\end{align}
All other aspects are kept the same as in the basic VQ-VAE (Sec.~\ref{sec:vq_vae}).
We refer to this approach as \textit{Kohonen-VAE}.

\begin{remark}[Relation to EMA-VQ]
If the neighbourhood is reduced to zero (see, Remark \ref{remark:k-means}),
this approach falls back to the standard EMA-VQ VAEs (Sec.~\ref{sec:vq_vae}).
\end{remark}

\subsection{Initialisation \& Updates of EMAs}
\label{sec:init}
\textbf{Initialisation.} Both the basic EMA-VQ (Sec.~\ref{sec:vq_vae}) and KSOM (Sec.~\ref{sec:k-vae})
require to initialise two EMAs: $\vm_k^{(0)}$ (Eq.~\ref{eq:vq_vae_sum} and \ref{eq:numerator_kvae}) and $N_k^{(0)}$ (Eq.~\ref{eq:denominator_vq_vae} and \ref{eq:denominator_kvae}).
This is an important detail which is omitted in the common description of VQ-VAEs.
Standard public implementations of VQ-VAEs, including the official one by van den Oord et al.~\cite{OordVK17},
initialise $\vm_k^{(0)}$ by a random Gaussian vector, and $N_k^{(0)}$ by $0$.
The latter is problematic for the following reason.

In fact, standard implementations apply the updates of Eqs.~\ref{eq:vq_vae_sum}-\ref{eq:vq_vae_update} to all clusters including those that have no members in the batch---later, we show that this is another important detail for EMA-VQ.
Since $N_k^{(0)}$ is initialised with $0$, $N_k^{(1)}$ remains $0$ at the first iteration for the clusters with no members in the first batch.
To avoid division by zero, smoothing over counts $(N_1^{(t)}, ..., N_K^{(t)})$---typically additive smoothing; see, e.g., \cite{chen1999empirical}---is applied to obtain smoothed counts $(\tilde{N}_1^{(t)}, ..., \tilde{N}_K^{(t)})$
such that $N_k^{(1)}=0$ becomes $\tilde{N}_k^{(1)} = \epsilon$ where typically $\epsilon \sim 10^{-5}$.
While this allows to avoid division by zero, the resulting multiplication of $\vm_k^{(1)}$ by $1 / \epsilon \sim 10^{5}$ in Eq.~\ref{eq:vq_vae_update} also seems unreasonable.
Our experiments (Sec.~\ref{sec:exp_init}) show that this effectively results in certain sub-optimality in training.
Instead, we propose $N_k^{(0)} = 1$ initialisation, which is consistent with non-zero initialisation of $\vm_k^{(0)}$.\looseness=-1

\textbf{Updating EMAs of Clusters without Members.}
Given the update equations above (Eqs.~\ref{eq:vq_vae_sum}-\ref{eq:vq_vae_update} and \ref{eq:numerator_kvae}-\ref{eq:quotient_kvae}), there remains the question whether the EMAs of clusters that have no members in the batch should be updated.
As mentioned already in the previous paragraph, standard implementations \textit{update all} EMAs.
We conduct the corresponding ablation study in Sec.~\ref{sec:exp_init}, and empirically show that, indeed, it is crucial to update all EMAs in the baseline EMA-VQ algorithm used in VQ-VAEs to achieve optimal performance.

Next, we'll also show that, unlike the standard EMA-VQ which is sensitive to these configurations, KSOM is robust, and performs well under any configurations.

\section{Experiments}
The goal of our experiments is to revisit the properties of KSOM when integrated into VQ-VAEs.
In particular, we demonstrate its robustness, and analyse learned representations.
Before that, we start with showing the sensitivity of the standard EMA-VQ w.r.t.~various configuration details.

\subsection{Sensitivity of the baseline EMA-VQ}
\label{sec:exp_init}
We first present two sets of experiments for the baseline EMA-VQ, which reveal its sensitivity to initialisation and update schemes of EMAs (discussed in Sec.~\ref{sec:init}).

\begin{figure}[h]
    \begin{center}
        \includegraphics[width=.5\columnwidth]{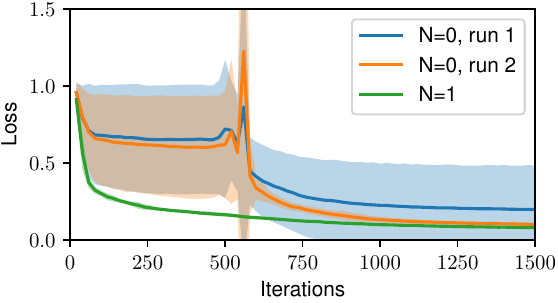}
        \caption{Evolution of validation reconstruction loss on CIFAR-10 for baseline EMA-VQ with different initialisations $N_k^{(0)}$ in Eq.~\ref{eq:denominator}.}
        \label{fig:fix_init}
    \end{center}
\end{figure}

\textbf{Improving Initialisation.}
We start with evaluating $N_k^{(0)}=1$ initialisation (instead of the standard $N_k^{(0)}=0$) discussed in Sec.~\ref{sec:init}.
Figure \ref{fig:fix_init} shows the evolution of the validation reconstruction loss of VQ-VAEs trained with EMA-VQ on CIFAR-10 \citep{krizhevsky} for two runs of 10 seeds each with $N_k^{(0)}=0$ (denoted by \texttt{N=0}/\texttt{run1} and \texttt{N=0}/\texttt{run2}), and one run of 10 seeds with $N_k^{(0)}=1$ (denoted by \texttt{N=1}).
In both runs with \texttt{N=0},
we observe a plateau at the beginning of training.
The variability of the results is also high: the performance of one of them (\texttt{N=0}/\texttt{run1}; the \textit{blue} curve) remains above that of the \texttt{N=1} case, even after the plateau, while the other one (\texttt{N=0}/\texttt{run2}, the \textit{orange} curve) successfully reaches the performance of \texttt{N=1}.
The final/best validation reconstruction losses ($1e^{-3}$) achieved by the respective configurations are: $148.7 \pm 299.7$ vs.~$52.1 \pm 0.6$.
In contrast, such a variability was not observed with $N_k^{(0)}=1$.

\textbf{Updating Clusters without Members.}
Now we also evaluate the effect of updating EMAs for the clusters that have no members in the batch.
Table \ref{tab:ablate_baseline} shows the corresponding results.
In all cases (with or without $N_k^{(0)}=1$ initialisation),
updating all clusters including those that have no member is crucial for good performance of the baseline EMA-VQ.

\begin{table}[h]
    \centering
    \caption{Sensitivity of the baseline EMA-VQ. VQ-VAEs trained with EMA-VQ on CIFAR-10. ``$N_k^{(0)}$'' indicates its initialisation with 0 or 1. ``Update-0'' indicates whether to update clusters that have 0 members in the batch. ``\# Steps'' is the number of training steps needed to achieve +10\% of the final loss. Mean and std are computed using 10 seeds. For the $N_k^{(0)}=0$ case, there is a high variability among results even with 10 seeds, as we report in Sec.~\ref{sec:exp_init}: here, we report the result obtained using 10 \textit{good} seeds.}
    \label{tab:ablate_baseline}
\begin{tabular}{ccrr}
\toprule
 $N_k^{(0)}$ & Update-0 & Loss ($1e^{-3}$)& \multicolumn{1}{c}{\# Steps ($1e^{3}$)} \\ \midrule
 0  & No & 148.8 $\pm$ 11.0 & 13.5 $\pm$ 0.7\\
 1 & No & 68.3 $\pm$ \; 0.8 & 16.3 $\pm$ 1.4\\ \midrule
 0  & Yes & 52.1  $\pm$ 0.6 & 6.8 $\pm$ 0.6 \\
 1 & Yes & \textBF{51.9} $\pm$ 0.2 & \textBF{5.4} $\pm$ 0.5 \\
\bottomrule
\end{tabular}
\end{table}

\subsection{Reconstruction Performance and Convergence Speed}

We evaluate the reconstruction performance and convergence speed of VQ-VAEs trained with KSOM
using three datasets: CIFAR-10 \citep{krizhevsky}, ImageNet \citep{DengDSLL009}, and a mixture of CelebA-HQ \citep{KarrasALL18} and Animal Faces HQ (AFHQ; \cite{ChoiUYH20}).
We use the basic VQ-VAE architecture \citep{OordVK17} for \mbox{CIFAR-10} and its extension VQ-VAE-2 \citep{RazaviOV19} for ImageNet and CelebA-HQ/AFHQ without architectural modifications. Further experimental details can be found in Appendix \ref{app:detail}\looseness=-1

\begin{table}[h]
\setlength{\tabcolsep}{0.5em}
    \centering
    \caption{Validation reconstruction loss and number of steps needed to achieve +10\% and +20\% of the final loss for various Kohonen-VAEs. ``None'' in column ``Neighbours'' corresponds to our \textit{well-configured} EMA-VQ (Sec.~\ref{sec:exp_init}; ``$N_k^{(0)}=1$'' and ``Update-0, Yes'' in Table \ref{tab:robustness_main}). 
    2D grid is used for both ``Hard'' and ``Gaussian'' cases. Mean and standard deviations are computed with 10 seeds for CIFAR-10 and 5 seeds each for ImageNet and CelebA-HQ/AFHQ.
    Image resolution is 32x32 for CIFAR-10 and 256x256 for ImageNet and CelebA-HQ/AFHQ. 
    }
    \label{tab:perf_optimal_baseline}
\begin{tabular}{llccc}
\toprule
 & &  & \multicolumn{2}{c}{\# Steps ($1e^{3}$)}  \\
 \cmidrule(lr){4-5}
Dataset & Neighbours & Loss ($1e^{-3}$) & +10\% & +20\% \\
\midrule
  & None (EMA-VQ) & 51.9 $\pm$ 0.2 & 5.4 $\pm$ 0.5 & 3.7 $\pm$ 0.5  \\
CIFAR-10 &Hard (KSOM) & 52.1 $\pm$ 0.2 &  \textBF{5.2} $\pm$ 0.6 &  \textBF{2.5} $\pm$ 0.5 \\
& Gaussian (KSOM)& \textBF{51.8} $\pm$ 0.2 &  \textBF{5.2} $\pm$ 0.6 & 3.0 $\pm$ 0.0  \\
\midrule
& None (EMA-VQ) & \textBF{23.0} $\pm$ 0.4 & 15.2 $\pm$ 1.6 & 8.4 $\pm$ 0.6  \\
ImageNet & Hard (KSOM) &  23.5 $\pm$ 0.4 &  \textBF{13.6} $\pm$ 2.2 &  \textBF{7.4} $\pm$ 0.9   \\
& Gaussian (KSOM) & 23.2 $\pm$ 0.4 & 14.0 $\pm$ 1.4 & 7.6 $\pm$ 0.6  \\
\midrule
& None (EMA-VQ) &  1.86 $\pm$ 0.10 & \textBF{17.0} $\pm$ 1.0 & 14.1 $\pm$ 1.4  \\
CelebA-HQ/AFHQ & Hard (KSOM) & \textBF{1.73} $\pm$ 0.01 &  \textBF{17.0} $\pm$ 1.6 &  \textBF{10.8} $\pm$ 1.3  \\
& Gaussian (KSOM) & 1.75 $\pm$ 0.03  & 17.8 $\pm$ 1.8 & 11.8 $\pm$ 1.3 \\
\bottomrule
\end{tabular}
\end{table}

\textbf{Comparison to Optimised EMA-VQ.}
We first compare models trained with KSOM with those trained using carefully configured EMA-VQ (Sec.~\ref{sec:exp_init}).
Table \ref{tab:perf_optimal_baseline} summarises the results.
We first observe that all methods achieve a similar validation reconstruction loss,
with slight improvements obtained by KSOM over the baseline on CelebA-HQ/AFHQ.
To compare the ``speed of convergence,'' we measure the number of steps needed by each algorithm to achieve $+10\%$ and $+20\%$ of their final performance.
Here ``steps'' correspond to the number of updates, and the batch size is the same for all methods.
We observe that, indeed, KSOM tends to be faster than the basic EMA-VQ at the beginning of training, as can be seen in the column $+20\%$ (especially for the hard variant on CelebA-HQ/AFHQ).
However, the baseline catches up later, and the difference becomes rather marginal at the $+10\%$ threshold: the corresponding speed up by KSOM is less than 5\% relative
compared to carefully configured EMA-VQ.
In what follows, we show that KSOM is much more robust than EMA-VQ, and performs well under all configurations, including those that are sub-optimal for EMA-VQ.\looseness=-1

\begin{table}[h]
 \setlength{\tabcolsep}{0.6em}
    \centering
    \caption{
Validation reconstruction loss and number of steps needed to achieve +10\% of the final loss (``\# Steps''), showing the \textbf{robustness of KSOM} w.r.t.~configurations that are sub-optimal for EMA-VQ. 
    ``$N_k^{(0)}$'' indicates its initialisation: 0 or 1. ``Update-0'' denotes whether to update clusters that have 0 members in the batch.
    Here we report the \textit{hard} variant (results are similar for \textit{Gaussian}).
    }
    \label{tab:robustness_main}
\begin{tabular}{llccrc}
\toprule
Dataset & Method & $N_k^{(0)}$ & Update-0 & \multicolumn{1}{c}{Loss ($1e^{-3}$)} & \# Steps ($1e^{3}$)  \\ \midrule

\multirow{7}{*}{CIFAR-10}& EMA-VQ & \multirow{2}{*}{0}  & \multirow{2}{*}{No} & 148.8 $\pm$ 11.0 & 13.5 $\pm$ 0.7  \\
&KSOM &   &  & \textBF{51.8} $\pm$ \: 0.2 & \, \textBF{5.2} $\pm$ 0.6   \\
\cmidrule(lr){2-6} 
&EMA-VQ &  \multirow{2}{*}{1} & \multirow{2}{*}{No} & 68.3 $\pm$ 0.8 & 16.3 $\pm$ 1.4   \\ 
&KSOM &   &  & \textBF{52.1} $\pm$ 0.3 & \,\,\,\textBF{4.8} $\pm$  0.9    \\
\cmidrule(lr){2-6} 
&EMA-VQ &  \multirow{2}{*}{0}  & \multirow{2}{*}{Yes} & 52.1  $\pm$ 0.6 & \, 6.8 $\pm$ 0.6  \\ 
&KSOM &    & & \textBF{51.9} $\pm$ 0.2 &  \,\,\,\textBF{4.8} $\pm$ 0.6 \\
\midrule
\midrule
\multirow{7}{*}{ImageNet}&EMA-VQ & \multirow{2}{*}{0}  & \multirow{2}{*}{No} &  44.7 $\pm$ 2.6 & 13.5 $\pm$ 0.7  \\
&KSOM &   &  & \textBF{24.4} $\pm$  1.3 & \textBF{12.6} $\pm$ 0.9 \\ 
\cmidrule(lr){2-6} 
&EMA-VQ &  \multirow{2}{*}{1} & \multirow{2}{*}{No} & 27.9 $\pm$ 1.0 & 21.2 $\pm$ 1.5    \\ 
&KSOM &   &  & \textBF{24.1} $\pm$ 1.3 & \textBF{12.2} $\pm$ 1.3      \\
\cmidrule(lr){2-6} 
&EMA-VQ &  \multirow{2}{*}{0}  & \multirow{2}{*}{Yes} & \textBF{23.6}  $\pm$ 0.9 & 15.4 $\pm$ 2.7   \\ 
&KSOM &    & &  23.7 $\pm$ 0.7 & \textBF{14.2} $\pm$ 1.6 \\
\midrule
\midrule
\multirow{7}{*}{CelebA-HQ/AFHQ}&EMA-VQ & \multirow{2}{*}{0}  & \multirow{2}{*}{No} & $3.73 \pm 0.12$  & $21.2 \pm 3.9$  \\
&KSOM &   &  & \textBF{1.74}\,$\pm$ 0.04  & \textBF{17.6} $\pm$ 3.6  \\ 
\cmidrule(lr){2-6} 
&EMA-VQ &  \multirow{2}{*}{1} & \multirow{2}{*}{No} &  $3.01 \pm 0.28$  & $22.6 \pm 0.5$  \\ 
&KSOM &   &  & \textBF{1.74}\,$\pm$ 0.02  & \textBF{16.6} $\pm$ 1.5   \\
\cmidrule(lr){2-6} 
&EMA-VQ &  \multirow{2}{*}{0}  & \multirow{2}{*}{Yes} & $1.80 \pm 0.10$  & $17.6 \pm 1.5$  \\ 
&KSOM &    & & \,\,\,\textBF{1.72}\,$\pm$ 0.01  & \textBF{16.8} $\pm$ 1.9 \\

\bottomrule
\end{tabular}
\end{table}

\textbf{Robustness of KSOM against EMA-VQ Issues.}
Above we report that the performance gain (both in speed and reconstruction quality) by KSOM is rather marginal compared to our carefully configured EMA-VQ baseline obtained in Sec.~\ref{sec:exp_init}.
Here we compare the two approaches under various configurations.
Table \ref{tab:robustness_main} shows the results.
We observe that KSOM is remarkably robust: in all configurations, including those that are sub-optimal for EMA-VQ,
KSOM achieves the same best validation loss as in the optimal configuration (Table \ref{tab:perf_optimal_baseline}).
KSOM's neighbourhood updating scheme naturally fixes the problematic cases of the original EMA-VQ above.
In these configurations, KSOM also generally converges faster than the baseline EMA-VQ.
These results hold for both VQ-VAEs trained on CIFAR-10 and
for VQ-VAE-2s trained on ImageNet and CelebA-HQ/AFHQ. In addition, in Appendix \ref{app:codebook}, we show that KSOM also tends to improve the codebook utilisation.

\subsection{Topologically Ordered Discrete Representations}
Finally, we analyse the discrete representations learned by KSOM.
We show that they are ``topologically'' ordered on the grid of indices, and consequently, reconstructed images remain close to the original ones even when we slightly shift their latent representations in the discrete index space.
\vspace{2mm}

\begin{figure}[h]
\begin{center}
	\subfloat[EMA-VQ]{
		\includegraphics[width=.49\linewidth]{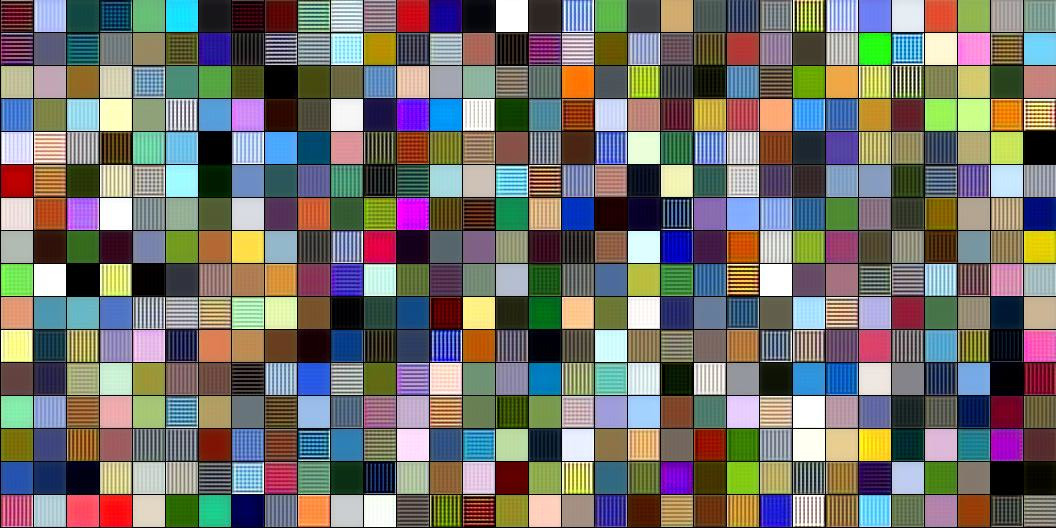}
	}
	\subfloat[KSOM]{
		\includegraphics[width=.49\linewidth]{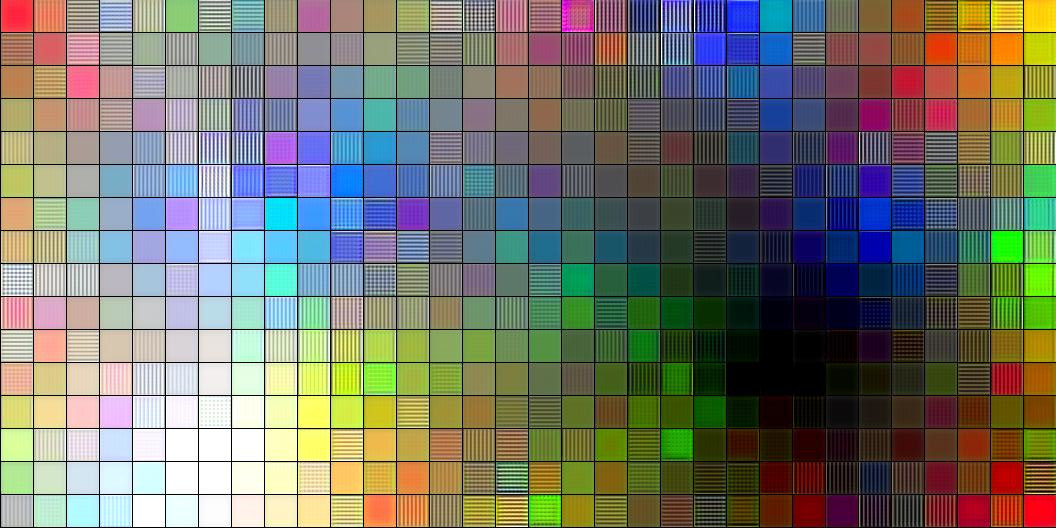}
	}
	\caption{A visualization of codebook ($K=512$) of VQ-VAEs trained on CIFAR10 with (a) EMA-VQ or (b) KSOM. 
 The codebook of EMA-VQ obviously has no structure but serves as a reference.
 Similar visualisations for VQ-VAE-2 trained on ImageNet can be found in Appendix \ref{app:vis}.
 }
	\label{fig:codebook_vis}
\end{center}
\end{figure}

\textbf{Grid Visualisation.}
We first visualise how learned discrete representations are distributed over the grid.
Here we use a VQ-VAE trained with KSOM (2D with hard neighbourhoods) on CIFAR-10, and proceed as follows.
A discrete latent representation consists of $N$ integers (Sec.~\ref{sec:vq_vae}).
For each index in the codebook (corresponding to one of the nodes on the grid),
we create a discrete latent representation whose $N$ codes are all the same and equal to the corresponding index, and feed it to the decoder to obtain an output image.
This results in a grid of images shown in Figure \ref{fig:codebook_vis}.
We observe that each code seems to correspond to some colour,
we can effectively observe several local ``islands'' of colours which group colours that are visually close.

\begin{figure*}[h]

		\centering
        \includegraphics[width=.9\linewidth]{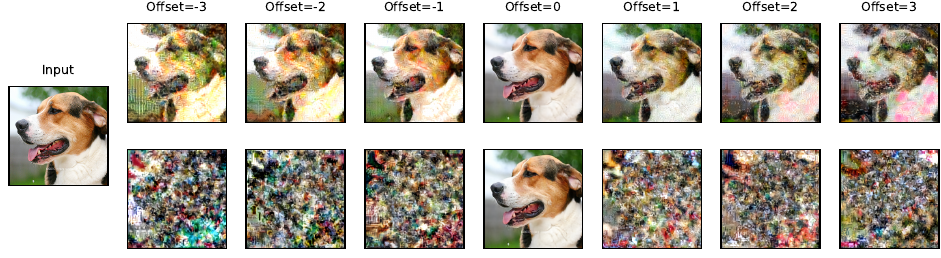}
 \vspace{5mm}
 \\
		\centering
        \includegraphics[width=.9\linewidth]{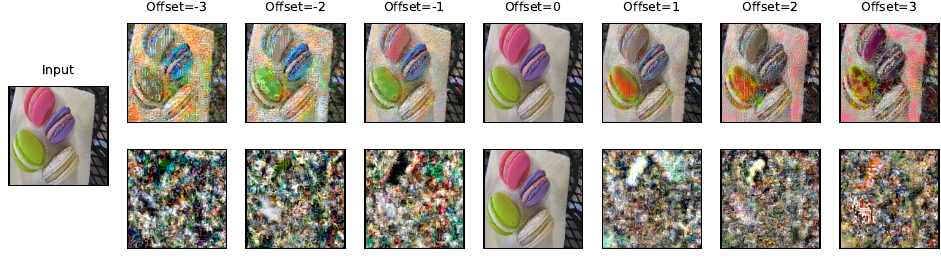}
	\caption{Effects of perturbations to the discrete latent code on reconstructed images. For each image, the \textbf{top row} shows the results for KSOM, and the \textbf{bottom row} shows those for EMA-VQ. ``Offset'' indicates the offset added to the indices of the latent representations.}
	\label{fig:shifting}
\end{figure*}

\textbf{Impact of Perturbation in the Discrete Latent Space.}
To further illustrate the presence of neighbourhoods developed by KSOM in the discrete latent space,
we show images obtained by perturbing the discrete latent code representing a proper image in the index space (by adding or subtracting an integer offset to each coordinates of the code indices).
Here we use VQ-VAE-2 trained on ImageNet with 2D KSOM with hard neighbourhoods.
The VQ-VAE-2 \citep{RazaviOV19} has two levels of discretisation: we shift all of them by the same offset on both $x$ and $y$ axes of the 2D grid for KSOM or directly shift the code indices for EMA-VQ.
Figure \ref{fig:shifting} shows the results.
Obviously, with the baseline VQ-VAE-2 trained with EMA-VQ, the output images become complete noises under such perturbations, even with an offset of one.
With the KSOM-trained representations,
there is a certain degree of continuity in the space of indices (as illustrated in Figure \ref{fig:codebook_vis}): the output images preserve the original contents though they become noisier as the offset increases.
This illustrates the neighbourhoods learned by KSOM.\looseness=-1

\section{Discussion}
\label{sec:discussion}

\textbf{Recommendations.}
Our first recommendation for any EMA-VQ implementation is to modify
$N_k^{(0)}=0$ to $N_k^{(0)}=1$ (Sec.~\ref{sec:exp_init}).
For a more robust solution, we recommend using KSOM.
Extending the standard implementation of EMA-VQ (Sec.~\ref{sec:vq_vae})
to KSOM (Sec.~\ref{sec:k-vae}) is straightforward.
While we introduce one extra hyper-parameter, shrink step $\tau$ (Eqs.~\ref{eq:shrink}-\ref{eq:gauss}), we found $\tau=0.1$ to perform well across all tasks.
Regarding the model variations, we recommend using the 2D variant with hard neighbourhoods.
Virtually, any VQ implementation should benefit from these modifications.
\vspace{2mm}

\textbf{Related Work.}
The most related work is Fortuin et al.~\cite{FortuinHLSR19}'s \textit{SOM-VAE}
which is also inspired by KSOM.
The core difference between the SOM-VAE and our approach is that Fortuin et al.~\cite{FortuinHLSR19} train the codebook weights by gradient descent:
all neighbour weight vectors are updated to be close to the encoder output
(see the last paragraph of Sec.~\ref{sec:vq_vae}).
While this is indeed inspired by Kohonen's neighbourhoods, none of Kohonen's learning rules (Sec.~\ref{sec:background}) is used.\footnote{Empirical results comparing KSOM and SOM-VAE is provided in Appendix \ref{app:comp_grad}. We find that the final reconstruction loss is similar
for the EMA baseline, SOM-VAE, and KSOM, but KSOM converges the fastest
(significantly faster than SOM-VAE). We also confirm that the EMA-based
codebook learning outperforms the gradient-based one as noted by
the original authors of VQ-VAE. We focus on KSOM because it is the natural extension of EMA which is recommended over the gradient-based variant (SOM-VAE is a natural extension of the gradient-based variant).\looseness=-1}
Also, the main focus of \cite{FortuinHLSR19} is on modelling and interpreting time series.
In fact, SOM-VAEs are extended by Manduchi et al.~\cite{ManduchiHFVRF21} for further applications to healthcare. Another type of VAEs exhibiting topographic properties can also be found in \citep{KellerW21}.\looseness=-1

There are also several other works which attempt to improve the VQ algorithm. 
For example, Zeghidour et al.~\cite{zeghidour2021soundstream} propose to initialise all codebook weights using examples in the first batch.
Lee et al.~\cite{lee2022autoregressive} propose residual VQ that iteratively performs VQ for improved reconstruction quality (note that \cite{lee2022autoregressive} use EMA-VQ which can be directly replaced by KSOM).

\textbf{Applications of VQ.}
Following the VQ-VAE of \cite{OordVK17},
VQ has become popular across various modalities and applications.
Besides the standard use case for high-dimensional data such as images, audio, and video,
VQ has been also used for texts. For example, Kaiser et al.~and Roy et al.~\cite{KaiserBRVPUS18, roy2018towards} downsample target sentences via VQ to speed up decoding in machine translation.
Liu et al.~\cite{liu2022learning} explore VQ to improve multi-lingual translation.
Ozair et al.~\cite{OzairLRAOV21} applies VQ to model-based reinforcement learning/planning by quantising state-action sequences.
Many text-to-image generation systems also include VQ components, e.g., Ramesh et al.~\cite{RameshPGGVRCS21} make use of discrete VAEs with Gumbel-softmax \citep{JangGP17}),
and Yu et al.~\cite{yu2022scaling} build upon VQ-GAN-2 \citep{YuLKZPQKXBW22}.
Discrete representation learning is also motivated by out-of-distribution generalisation in certain tasks \citep{LiuLKGSMB21, liu2022adaptive, trauble2022discrete}.
\vspace{2mm}

\textbf{Differentiable Relaxation of KSOM.}
While working very well in practice,
the use of the straight-through estimator (Sec.~\ref{sec:vq_vae}) to pass the gradients through the quantisation operation seems sub-optimal at first.
For example, Agustsson et al.~\cite{AgustssonMTCTBG17} show the possibility to perform discrete representation learning in fully differentiable NNs by using softmax with temperature annealing.
In fact, we can also naturally derive a differentiable version of KSOM by replacing $\hardmax$ in Eq.~\ref{eq:find_max_mat} (presented in Sec.~\ref{sec:background}) by the softmax function.
However, in our preliminary experiments,
none of our differentiable variants with temperature annealing obtained a successful model that achieves a good reconstruction loss when the softmax is discretised for testing.
\vspace{2mm}

\textbf{Semantic Codebook.}
While we demonstrate the emergence of neighbourhoods in the discrete space of codebook indices learned by KSOM,
the features encoded by them remain low-level (mostly colours).
While learning of more ``semantic'' discrete codes is out of scope here,
we expect KSOM with such representations to be even more interesting, as it may potentially enable interpolation in the discrete latent space.
\vspace{2mm}

\textbf{Other Variations of KSOM.}
Finally, there are several extensions of KSOM, e.g., \textit{neural gas} \citep{martinetz1991neural, Fritzke94}.
Hopefully, our work will inspire further research on improving discrete representation learning in modern NNs through KSOM variations or other algorithms going beyond EMA-VQ.

\section{Conclusion}
We revisit the learning rule of Kohonen's Self-Organising Maps (KSOM) 
as the vector quantisation (VQ) algorithm for discrete representation learning in neural networks.
KSOM is a generalisation of the exponential moving-average based VQ algorithm (EMA-VQ) commonly used in VQ-VAEs.
We empirically demonstrate that, unlike the standard EMA-VQ, KSOM is robust w.r.t.~initialisation and EMA update schemes.
Our recipes can easily be integrated into existing code for VQ-VAEs.
In addition, we show that discrete representations learned by KSOM effectively develop topological structures.
We provide illustrations of the learned neighbourhoods in the image domain.

\begin{credits}
\subsubsection{\ackname}
This research was partially funded by ERC Advanced grant no: 742870, project AlgoRNN,
and by Swiss National Science Foundation grant no: 200021\_192356, project NEUSYM.
We are thankful for hardware donations from NVIDIA and IBM.
The resources used for this work were partially provided by Swiss National Supercomputing Centre (CSCS) project s1145 and s1154.
\end{credits}

\bibliography{paper}

\begin{thebibliography}{10}
\providecommand{\url}[1]{\texttt{#1}}
\providecommand{\urlprefix}{URL }
\providecommand{\doi}[1]{https://doi.org/#1}

\bibitem{AgustssonMTCTBG17}
Agustsson, E., Mentzer, F., Tschannen, M., Cavigelli, L., Timofte, R., Benini, L., Gool, L.V.: Soft-to-hard vector quantization for end-to-end learning compressible representations. In: Proc. Advances in Neural Information Processing Systems (NIPS). pp. 1141--1151 (Dec 2017)

\bibitem{amari1980topographic}
Amari, S.I.: Topographic organization of nerve fields. Bulletin of Mathematical Biology  \textbf{42}(3),  339--364 (1980)

\bibitem{BaevskiSA20}
Baevski, A., Schneider, S., Auli, M.: vq-wav2vec: Self-supervised learning of discrete speech representations. In: Int. Conf. on Learning Representations (ICLR). Virtual only (Apr 2020)

\bibitem{bengio2013estimating}
Bengio, Y., L{\'e}onard, N., Courville, A.: Estimating or propagating gradients through stochastic neurons for conditional computation. Preprint arXiv:1308.3432  (2013)

\bibitem{borsos2022audiolm}
Borsos, Z., Marinier, R., Vincent, D., Kharitonov, E., Pietquin, O., Sharifi, M., Teboul, O., Grangier, D., Tagliasacchi, M., Zeghidour, N.: Audio{LM}: a language modeling approach to audio generation. Preprint arXiv:2209.03143  (2022)

\bibitem{chen1999empirical}
Chen, S.F., Goodman, J.: An empirical study of smoothing techniques for language modeling. Computer Speech \& Language  \textbf{13}(4),  359--393 (1999)

\bibitem{ChoiUYH20}
Choi, Y., Uh, Y., Yoo, J., Ha, J.: Star{GAN} v2: Diverse image synthesis for multiple domains. In: Proc. {IEEE} Conf. on Computer Vision and Pattern Recognition (CVPR). pp. 8185--8194. Virtual only (Jun 2020)

\bibitem{constantinescu2016organizing}
Constantinescu, A.O., O’Reilly, J.X., Behrens, T.E.: Organizing conceptual knowledge in humans with a gridlike code. Science  \textbf{352}(6292),  1464--1468 (2016)

\bibitem{cottrell2018self}
Cottrell, M., Olteanu, M., Rossi, F., Villa-Vialaneix, N.: Self-organizing maps, theory and applications. Revista de Investigacion Operacional  \textbf{39}(1),  1--22 (2018)

\bibitem{ctl2022}
Csord{\'a}s, R., Irie, K., Schmidhuber, J.: {CTL}++: Evaluating generalization on never-seen compositional patterns of known functions, and compatibility of neural representations. In: Proc. Conf. on Empirical Methods in Natural Language Processing (EMNLP). Abu Dhabi, UAE (Dec 2022)

\bibitem{de2004use}
De~Bodt, E., Cottrell, M., Letremy, P., Verleysen, M.: On the use of self-organizing maps to accelerate vector quantization. Neurocomputing  \textbf{56},  187--203 (2004)

\bibitem{DengDSLL009}
Deng, J., Dong, W., Socher, R., Li, L., Li, K., Fei{-}Fei, L.: Image{N}et: {A} large-scale hierarchical image database. In: Proc. {IEEE} Conf. on Computer Vision and Pattern Recognition (CVPR). pp. 248--255. Miami, Florida, {USA} (Jun 2009)

\bibitem{dhariwal2020jukebox}
Dhariwal, P., Jun, H., Payne, C., Kim, J.W., Radford, A., Sutskever, I.: Jukebox: A generative model for music. Preprint arXiv:2005.00341  (2020)

\bibitem{EsserRO21}
Esser, P., Rombach, R., Ommer, B.: Taming transformers for high-resolution image synthesis. In: Proc. {IEEE} Conf. on Computer Vision and Pattern Recognition (CVPR). pp. 12873--12883. Virtual only (Jun 2021)

\bibitem{FortuinHLSR19}
Fortuin, V., H{\"{u}}ser, M., Locatello, F., Strathmann, H., R{\"{a}}tsch, G.: {SOM-VAE:} interpretable discrete representation learning on time series. In: Int. Conf. on Learning Representations (ICLR). New Orleans, LA, USA (May 2019)

\bibitem{Fritzke94}
Fritzke, B.: A growing neural gas network learns topologies. In: Proc. Advances in Neural Information Processing Systems (NIPS). pp. 625--632. Denver, CO, USA (1994)

\bibitem{hebb1949organization}
Hebb, D.O.: The organization of behavior; a neuropsycholocigal theory. A Wiley Book in Clinical Psychology  \textbf{62}, ~78 (1949)

\bibitem{hinton2012neural}
Hinton, G.: Neural networks for machine learning. Coursera, video lectures  (2012)

\bibitem{HuMTMS17}
Hu, W., Miyato, T., Tokui, S., Matsumoto, E., Sugiyama, M.: Learning discrete representations via information maximizing self-augmented training. In: Proc. Int. Conf. on Machine Learning (ICML). pp. 1558--1567. Sydney, Australia (Aug 2017)

\bibitem{hupkes2018learning}
Hupkes, D., Singh, A., Korrel, K., Kruszewski, G., Bruni, E.: Learning compositionally through attentive guidance. In: Proc. Int. Conf. on Computational Linguistics and Intelligent Text Processing. La Rochelle, France (Apr 2019)

\bibitem{JangGP17}
Jang, E., Gu, S., Poole, B.: Categorical reparameterization with gumbel-softmax. In: Int. Conf. on Learning Representations (ICLR). Toulon, France (Apr 2017)

\bibitem{KaiserBRVPUS18}
Kaiser, L., Bengio, S., Roy, A., Vaswani, A., Parmar, N., Uszkoreit, J., Shazeer, N.: Fast decoding in sequence models using discrete latent variables. In: Proc. Int. Conf. on Machine Learning (ICML). pp. 2395--2404. Stockholm, Sweden (Jul 2018)

\bibitem{KarrasALL18}
Karras, T., Aila, T., Laine, S., Lehtinen, J.: Progressive growing of {GAN}s for improved quality, stability, and variation. In: Int. Conf. on Learning Representations (ICLR). Vancouver, Canada (Apr 2018)

\bibitem{KellerW21}
Keller, T.A., Welling, M.: Topographic vaes learn equivariant capsules. In: Proc. Advances in Neural Information Processing Systems (NeurIPS). pp. 28585--28597. Virtual only (Dec 2021)

\bibitem{KingmaW13}
Kingma, D.P., Welling, M.: Auto-encoding variational bayes. In: Int. Conf. on Learning Representations (ICLR). Banff, Canada (Apr 2014)

\bibitem{kohonen1982self}
Kohonen, T.: Self-organized formation of topologically correct feature maps. Biological cybernetics  \textbf{43}(1),  59--69 (1982)

\bibitem{kohonen1999comparison}
Kohonen, T.: Comparison of {SOM} point densities based on different criteria. Neural Computation  \textbf{11}(8),  2081--2095 (1999)

\bibitem{kohonen2001book}
Kohonen, T.: Self-organizing maps. Springer (the first edition published in 1995) (2001)

\bibitem{krizhevsky}
Krizhevsky, A.: Learning multiple layers of features from tiny images. Master's thesis, Computer Science Department, University of Toronto (2009)

\bibitem{lee2022autoregressive}
Lee, D., Kim, C., Kim, S., Cho, M., Han, W.S.: Autoregressive image generation using residual quantization. In: Proc. {IEEE} Conf. on Computer Vision and Pattern Recognition (CVPR). pp. 11523--11532. New Orleans, LA, USA (Jun 2022)

\bibitem{liska2018memorize}
Liska, A., Kruszewski, G., Baroni, M.: Memorize or generalize? searching for a compositional {RNN} in a haystack. In: AEGAP Workshop ICML. Stockholm, Sweden (July 2018)

\bibitem{liu2022learning}
Liu, D., Niehues, J.: Learning an artificial language for knowledge-sharing in multilingual translation. In: Proc. Conf. on Machine Translation (WMT). pp. 188--202. Abu Dhabi (December 2022)

\bibitem{liu2022adaptive}
Liu, D., Lamb, A., Ji, X., Notsawo, P., Mozer, M., Bengio, Y., Kawaguchi, K.: Adaptive discrete communication bottlenecks with dynamic vector quantization. Preprint arXiv:2202.01334  (2022)

\bibitem{LiuLKGSMB21}
Liu, D., Lamb, A., Kawaguchi, K., Goyal, A., Sun, C., Mozer, M.C., Bengio, Y.: Discrete-valued neural communication. In: Proc. Advances in Neural Information Processing Systems (NeurIPS). pp. 2109--2121. Virtual only (Dec 2021)

\bibitem{lloyd1982least}
Lloyd, S.: Least squares quantization in {PCM}. IEEE Transactions on Information Theory  \textbf{28}(2),  129--137 (1982)

\bibitem{macqueen1967classification}
MacQueen, J.: Classification and analysis of multivariate observations. In: Proc. Berkeley Symp. Math. Statist. Probability. pp. 281--297 (1967)

\bibitem{von1973self}
von~der Malsburg, C.: Self-organization of orientation sensitive cells in the striate cortex. Kybernetik  \textbf{14}(2),  85--100 (1973)

\bibitem{von1977label}
von~der Malsburg, C., Willshaw, D.J.: How to label nerve cells so that they can interconnect in an ordered fashion. Proc. the National Academy of Sciences  \textbf{74}(11),  5176--5178 (1977)

\bibitem{ManduchiHFVRF21}
Manduchi, L., H{\"{u}}ser, M., Faltys, M., Vogt, J.E., R{\"{a}}tsch, G., Fortuin, V.: {T-DPSOM:} an interpretable clustering method for unsupervised learning of patient health states. In: Proc. Conf. on Health, Inference, and Learning (CHIL). pp. 236--245. Virtual only (Apr 2021)

\bibitem{martinetz1991neural}
Martinetz, T., Schulten, K.: A ``neural-gas'' network learns topologies. In: Proc. Int. Conf. on Artificial Neural Networks (ICANN). Espoo, Finland (Jun 1991)

\bibitem{nasrabadi1988vector}
Nasrabadi, N.M., Feng, Y.: Vector quantization of images based upon the {K}ohonen self-organizing feature maps. In: Proc.~IEEE Int. Conf. on Neural Networks (ICNN). vol.~1, pp. 101--105 (1988)

\bibitem{oja1982simplified}
Oja, E.: Simplified neuron model as a principal component analyzer. Journal of mathematical biology  \textbf{15}(3),  267--273 (1982)

\bibitem{OordVK17}
van~den Oord, A., Vinyals, O., Kavukcuoglu, K.: Neural discrete representation learning. In: Proc. Advances in Neural Information Processing Systems (NIPS). pp. 6306--6315. Long Beach, CA (Dec 2017)

\bibitem{OzairLRAOV21}
Ozair, S., Li, Y., Razavi, A., Antonoglou, I., van~den Oord, A., Vinyals, O.: Vector quantized models for planning. In: Proc. Int. Conf. on Machine Learning (ICML). pp. 8302--8313. Virtual only (Jul 2021)

\bibitem{RameshPGGVRCS21}
Ramesh, A., Pavlov, M., Goh, G., Gray, S., Voss, C., Radford, A., Chen, M., Sutskever, I.: Zero-shot text-to-image generation. In: Proc. Int. Conf. on Machine Learning (ICML). vol.~139, pp. 8821--8831. Virtual only (Dec 2021)

\bibitem{RazaviOV19}
Razavi, A., van~den Oord, A., Vinyals, O.: Generating diverse high-fidelity images with {VQ-VAE-2}. In: Proc. Advances in Neural Information Processing Systems (NeurIPS). pp. 14837--14847. Vancouver, Canada (Dec 2019)

\bibitem{RombachBLEO22}
Rombach, R., Blattmann, A., Lorenz, D., Esser, P., Ommer, B.: High-resolution image synthesis with latent diffusion models. In: Proc. {IEEE} Conf. on Computer Vision and Pattern Recognition (CVPR). pp. 10674--10685. New Orleans, LA, USA (Jun 2022)

\bibitem{roy2018towards}
Roy, A., Vaswani, A., Parmar, N., Neelakantan, A.: Towards a better understanding of vector quantized autoencoders. OpenReview  (2018)

\bibitem{schlag2021linear}
Schlag, I., Irie, K., Schmidhuber, J.: Linear {T}ransformers are secretly fast weight programmers. In: Proc. Int. Conf. on Machine Learning (ICML). Virtual only (Jul 2021)

\bibitem{Schmidhuber:91fastweights}
Schmidhuber, J.: Learning to control fast-weight memories: An alternative to recurrent nets. Tech. Rep. FKI-147-91, Institut f\"{u}r Informatik, Technische Universit\"{a}t M\"{u}nchen (March 1991)

\bibitem{tirunagari2016visualisation}
Tirunagari, S., Bull, S., Kouchaki, S., Cooke, D., Poh, N.: Visualisation of survey responses using self-organising maps: a case study on diabetes self-care factors. In: Proc.~IEEE Symposium Series on Computational Intelligence (SSCI). pp.~1--6 (2016)

\bibitem{TjandraS020}
Tjandra, A., Sakti, S., Nakamura, S.: Transformer {VQ-VAE} for unsupervised unit discovery and speech synthesis: Zerospeech 2020 challenge. In: Proc. Interspeech. pp. 4851--4855. Virtual only (Oct 2020)

\bibitem{trauble2022discrete}
Tr{\"a}uble, F., Goyal, A., Rahaman, N., Mozer, M., Kawaguchi, K., Bengio, Y., Sch{\"o}lkopf, B.: Discrete key-value bottleneck. Preprint arXiv:2207.11240  (2022)

\bibitem{trafo}
Vaswani, A., Shazeer, N., Parmar, N., Uszkoreit, J., Jones, L., Gomez, A.N., Kaiser, {\L}., Polosukhin, I.: Attention is all you need. In: Proc. Advances in Neural Information Processing Systems (NIPS). pp. 5998--6008. Long Beach, {CA}, {USA} (Dec 2017)

\bibitem{walker2021predicting}
Walker, J., Razavi, A., Oord, A.v.d.: Predicting video with {VQVAE}. Preprint arXiv:2103.01950  (2021)

\bibitem{willshaw1976patterned}
Willshaw, D.J., von~der Malsburg, C.: How patterned neural connections can be set up by self-organization. Proceedings of the Royal Society of London. Series B. Biological Sciences  \textbf{194}(1117),  431--445 (1976)

\bibitem{willshaw1979marker}
Willshaw, D.J., von~der Malsburg, C.: A marker induction mechanism for the establishment of ordered neural mappings: its application to the retinotectal problem. Philosophical Transactions of the Royal Society of London. B, Biological Sciences  \textbf{287}(1021),  203--243 (1979)

\bibitem{yan2021videogpt}
Yan, W., Zhang, Y., Abbeel, P., Srinivas, A.: Video{GPT}: Video generation using vq-vae and transformers. Preprint arXiv:2104.10157  (2021)

\bibitem{yin2008self}
Yin, H.: The self-organizing maps: background, theories, extensions and applications. In: Computational intelligence: A compendium, pp. 715--762 (2008)

\bibitem{YuLKZPQKXBW22}
Yu, J., Li, X., Koh, J.Y., Zhang, H., Pang, R., Qin, J., Ku, A., Xu, Y., Baldridge, J., Wu, Y.: Vector-quantized image modeling with improved {VQGAN}. In: Int. Conf. on Learning Representations (ICLR). Virtual only (Apr 2022)

\bibitem{yu2022scaling}
Yu, J., Xu, Y., Koh, J.Y., Luong, T., Baid, G., Wang, Z., Vasudevan, V., Ku, A., Yang, Y., Ayan, B.K., et~al.: Scaling autoregressive models for content-rich text-to-image generation. Preprint arXiv:2206.10789  (2022)

\bibitem{zeghidour2021soundstream}
Zeghidour, N., Luebs, A., Omran, A., Skoglund, J., Tagliasacchi, M.: Soundstream: An end-to-end neural audio codec. IEEE/ACM Transactions on Audio, Speech, and Language Processing  \textbf{30},  495--507 (2021)

\end{thebibliography}
\bibliographystyle{splncs04}

\newpage

\section*{Appendix}

\section{Experimental Details}
\label{app:detail}
\textbf{Datasets.}
We use three datasets (one of them is a mixture of two standard datasets): CIFAR-10 \citep{krizhevsky}, ImageNet \citep{DengDSLL009}, and a mixture of CelebA-HQ \citep{KarrasALL18} and Animal Faces HQ (AFHQ \citep{ChoiUYH20}).
The number of images and the resolution we use are: 60\,K and 32x32, 1.3\,M and 256x256, 43\,K (the sum of 28\,K for CelebA-HQ and 15\,K for AFHQ) and 256x256, respectively.
For any further details, we refer the readers to the original references.

\textbf{Model Architectures.}
We use the basic configuration (including hyper-parameters) of VQ-VAE \citep{OordVK17} for \mbox{CIFAR-10} and that of VQ-VAE-2 \citep{RazaviOV19} for ImageNet and CelebA-HQ/AFHQ from the original papers.
The only change we introduce is the learning algorithm to train the codebook weights (Sec.~\ref{sec:k-vae}).
Again, for any further details on the model architectures, we refer the readers to the original references above.
The dimension of latent representation for VQ-VAE on CIFAR-10 is 8x8x64, i.e, $N= 8 \times 8 = 64$ and $d_{\text{emb}}=64$ using our notations of Sec.~\ref{sec:vq_vae}.
VQ-VAE-2 has two codebooks (``top'' and ``bottom'') with respective embedding dimensions of 32x32x64 ($N= 32 \times 32$) and 64x64x64 ($N= 64 \times 64$) with $d_{\text{emb}}=64$ for both.
We use the codebook size of $K=512$ everywhere for all datasets.

\textbf{Reference Baseline Implementations.}
We looked into several public implementations of the baseline VQ-VAEs, including
the original implementation of VQ-VAE/EMA-VQ:
\url{https://github.com/deepmind/sonnet/blob/v2/sonnet/src/nets/vqvae.py}, its PyTorch re-implementation: \url{https://github.com/zalandoresearch/pytorch-vq-vae} (see also:
\url{https://github.com/lucidrains/vector-quantize-pytorch}),
and VQ-VAE-2: \url{https://github.com/rosinality/vq-vae-2-pytorch}.
The GitHub link to our code can be found on the first page.

\section{Extra Experimental Results}
Here we provide several extra experimental results which we could not report in the main text due to space limitation.

\subsection{Codebook Utilisation}
\label{app:codebook}
\begin{figure}[t]
\begin{center}
	\subfloat[CIFAR-10, VQ-VAE]{
  \includegraphics[width=.45\linewidth]{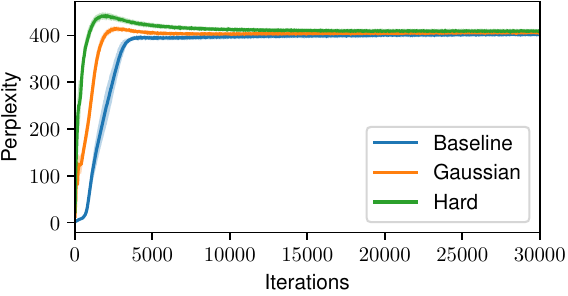}
        }
 \\
	\subfloat[ImageNet, VQ-VAE-2, Top]{
		\includegraphics[width=.45\linewidth]{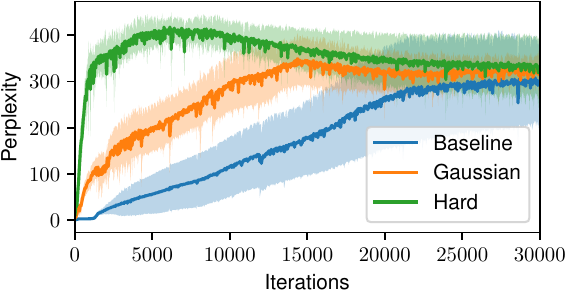}
	}
 \hspace{2mm}
        \subfloat[ImageNet, VQ-VAE-2, Bottom]{
		\includegraphics[width=.48\linewidth]{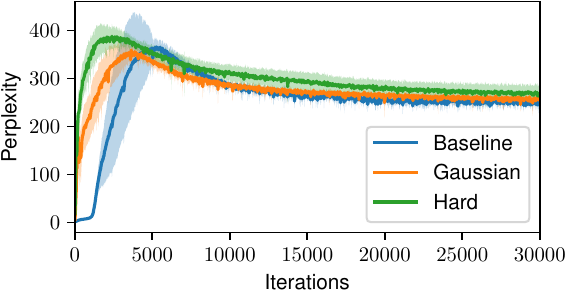}
	}
 \\
        \subfloat[CelebA-HQ/AFHQ, VQ-VAE-2, Top]{
  \includegraphics[width=.45\linewidth]{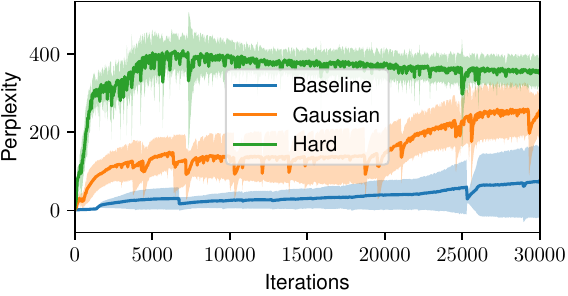}
	}
 \hspace{2mm}
        \subfloat[CelebA-HQ/AFHQ, VQ-VAE-2, Bottom]{
		\includegraphics[width=.48\linewidth]{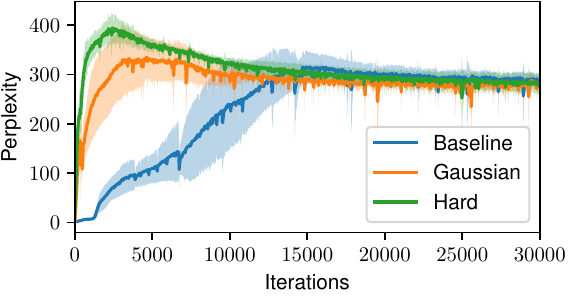}
	}
	\caption{Evolution of perplexity (codebook utilisation) as a function of training iterations. The codebook size is $K=512$ in all cases. ``Baseline'' is the standard EMA-VQ. ``Hard'' and ``Gaussian'' indicate the corresponding neighbourhood type for KSOM.}
	\label{fig:ppl}
 \end{center}
\end{figure}

Here we show that KSOM also helps boosting the codebook utilisation of VQ-VAE and VQ-VAE-2.
As a measure of codebook utilisation, we compute \textit{perplexity} for each batch as
$\exp(\sum_{k=1}^K p(k) \log(p(k)))$ where $p(k) = C(k) / (N * B)$ with $C(k)$ denoting the number of times the code $k$ is used in the batch (with a batch size $B$ and the number of embeddings $N$; the same notation as in the main text).
Figure \ref{fig:ppl} shows the evolution of perplexity on all datasets (for VQ-VAE-2s, we show that for the two codebooks).
With the only exception of the ``bottom'' codebook of VQ-VAE-2 trained on CelebA-HQ/AFHQ, overall, we observe that the code utilisation of KSOM variants tends to be higher than that of the baseline EMA-VQ.
In particular, for the ``top'' codebook of VQ-VAE-2 trained on CelebA-HQ/AFHQ (Figure \ref{fig:ppl} (d)),
the hard KSOM's mean perplexity exceeds 300 while the baseline EMA-VQ's is below 100.

\subsection{Ablation Studies}
\label{app:ablation}
In the main text, we show that the performances of the hard and Gaussian variants are rather close (Table \ref{tab:perf_optimal_baseline}).
We also conduct ablation studies to compare 1D vs.~2D grids, and different values of shrinking step $\tau \in \{1, 0.1, 0.01\}$,
and observe that these different variations yield rather similar performance.
For the sake of completeness, Table \ref{tab:app_ablation} presents the corresponding results.
In terms of neighbourhoods, as expected, we find that the variants with small shrinking steps $\tau$ of $0.01$ or $0.1$ yield smoother codebook grids than those obtained with $\tau=1$.

\begin{table}[t]
\small
\setlength{\tabcolsep}{0.2em}
    \centering
    \caption{Ablation studies on the 1d/2D grid and shrinking step size $\tau$. Validation reconstruction loss and number of steps needed to achieve +10\% and +20\% of the final loss.
    Mean and standard deviations are computed with 10 seeds for CIFAR-10 and 5 seeds each for CelebA-HQ/AFHQ.
    Image resolution is 32x32 for CIFAR-10 and 256x256 for CelebA-HQ/AFHQ. 
    The best numbers for ``hard'' and ``Gaussian'' groups are highlighted in \textbf{bold} separately.}
    \label{tab:app_ablation}
\begin{tabular}{llcccccc}
\toprule
& & \multicolumn{3}{c}{CIFAR-10} &  \multicolumn{3}{c}{CelebA-HQ/AFHQ} \\
\cmidrule(lr){3-5}\cmidrule(lr){6-8}
&  &  & \multicolumn{2}{c}{\# Steps ($1e^{3}$)} &  & \multicolumn{2}{c}{\# Steps ($1e^{3}$)} \\
Neighbours & $\tau$ & Loss ($1e^{-3}$) & +10\% & +20\% & Loss ($1e^{-4}$) & +10\% & +20\% \\
    \midrule
Hard 1D & 0.1 & \textBF{52.0} $\pm$ 0.2  & $5.3 \pm 0.5$  & $3.1 \pm 0.3$  & $17.4 \pm 0.2$  & \textBF{15.8} $\pm$ 0.8  & \textBF{10.8} $\pm$ 1.5  \\
\midrule
Hard 2D & 1 & $52.1 \pm 0.4$  & \textBF{4.7} $\pm$ 0.7  & 2.6 $\pm$ 0.5  & $17.4 \pm 0.2$  & $18.0 \pm 1.6$  & $11.4 \pm 1.1$  \\
Hard 2D & 0.1 & \textBF{52.0} $\pm$ 0.3  & 4.9 $\pm$ 0.3  & \textBF{2.4} $\pm$ 0.5  & \textBF{17.3} $\pm$ 0.2  & $18.0 \pm 1.6$  & $12.4 \pm 1.1$  \\
Hard 2D & 0.01 & $52.1 \pm 0.3$  & $5.5 \pm 0.7$  & $3.0 \pm 0.0$  & $17.6 \pm 0.3$  & $17.0 \pm 2.9$  & $12.4 \pm 2.1$  \\
\midrule
Gaussian 1D & 0.1 & $51.9 \pm 0.4$  & $5.2 \pm 0.4$  & $3.1 \pm 0.3$  & $17.8 \pm 0.5$  & $18.0 \pm 2.7$  & \textBF{11.6} $\pm$ 1.3  \\
\midrule
Gaussian 2D & 1 & $51.9 \pm 0.2$  & $5.3 \pm 0.5$  & $3.3 \pm 0.5$  & $18.3 \pm 0.4$  & \textBF{16.4} $\pm$ 1.9  & 11.8 $\pm$ 1.8  \\
Gaussian 2D & 0.1 & $51.8 \pm 0.3$  & \textBF{5.1} $\pm$ 0.3 & \textBF{3.1} $\pm$ 0.3  & $17.6 \pm 0.2$  & $18.6 \pm 2.1$  & $13.2 \pm 0.8$  \\
Gaussian 2D & 0.01 & \textBF{51.7} $\pm$ 0.2  & $5.4 \pm 0.5$  & $3.6 \pm 0.5$  & \textBF{17.5} $\pm$ 0.5 & $18.4 \pm 2.6$  & $11.8 \pm 1.3$  \\
\bottomrule
\end{tabular}
\end{table}

\subsection{Comparison to Gradient-based Codebook Learning Methods}
\label{app:comp_grad}
Here we compare KSOM to gradient-based codebook learning methods, including SOM-VAE \citep{FortuinHLSR19}.
Table \ref{tab:comp_grad} shows the results.
The final reconstruction loss is similar for the EMA baseline, SOM-VAE, and KSOM, but KSOM converges the fastest (significantly faster than SOM-VAE). We also confirm that the EMA-based codebook learning outperforms the gradient-based variant (first row) as noted by the original authors of VQ-VAE (see, \url{https://twitter.com/avdnoord/status/1001853279649910784?lang=en}).
We focus on KSOM because it is the natural extension of EMA which is recommended over the gradient-based variant (while SOM-VAE is a natural extension of the gradient-based variant).

\begin{table}[t]
    \centering
    \caption{
Validation reconstruction loss and number of steps needed to achieve +10\% of the final loss (``\# Steps'') on \textbf{CIFAR-10}.}
    \label{tab:comp_grad}
\begin{tabular}{lrrr}
\toprule
Codebook Learning Algorithm & Loss ($1e^{-3}$) & \# Steps ($1e^{3}$) \\ \midrule
Gradient-based & 69.4 $\pm$ 0.8 & 14.3 $\pm$ 0.9 \\
EMA & 51.9 $\pm$ 0.2 & 5.4 $\pm$ 0.5 \\ \midrule
SOM-VAE (Gradient-based) & 51.7 $\pm$ 0.3 & 7.3 $\pm$ 0.5 \\
KSOM & 52.1 $\pm$ 0.2 & 5.2 $\pm$ 0.6 \\
\bottomrule
\end{tabular}
\end{table}

\subsection{More Visualisations for VQ-VAE-2}
\label{app:vis}

\begin{figure}[h]
\begin{center}
	\subfloat[Top=(0, 0)]{
		\includegraphics[width=.3\linewidth]{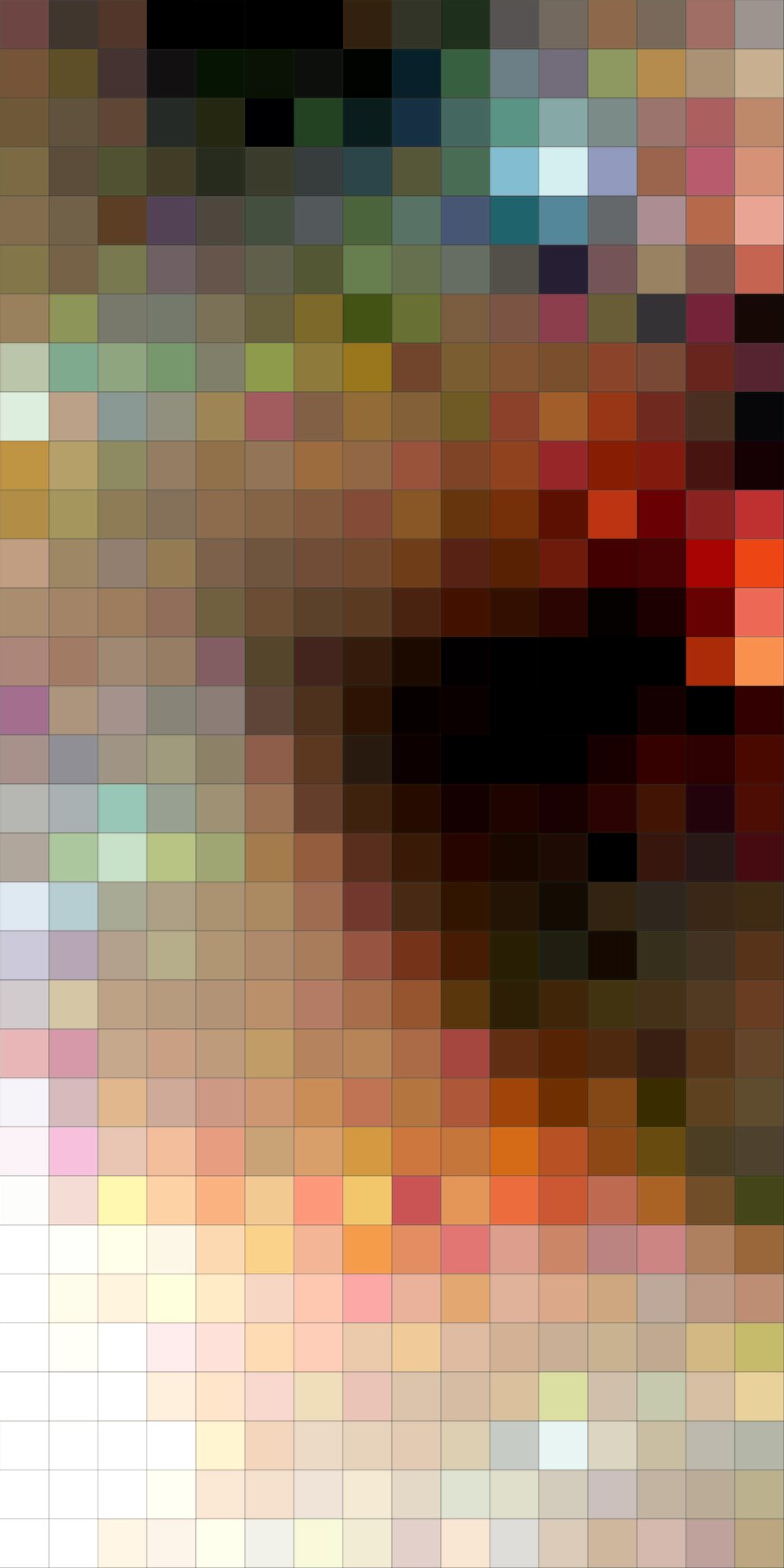}
	}
        \subfloat[Top=(18, 15)]{
		\includegraphics[width=.3\linewidth]{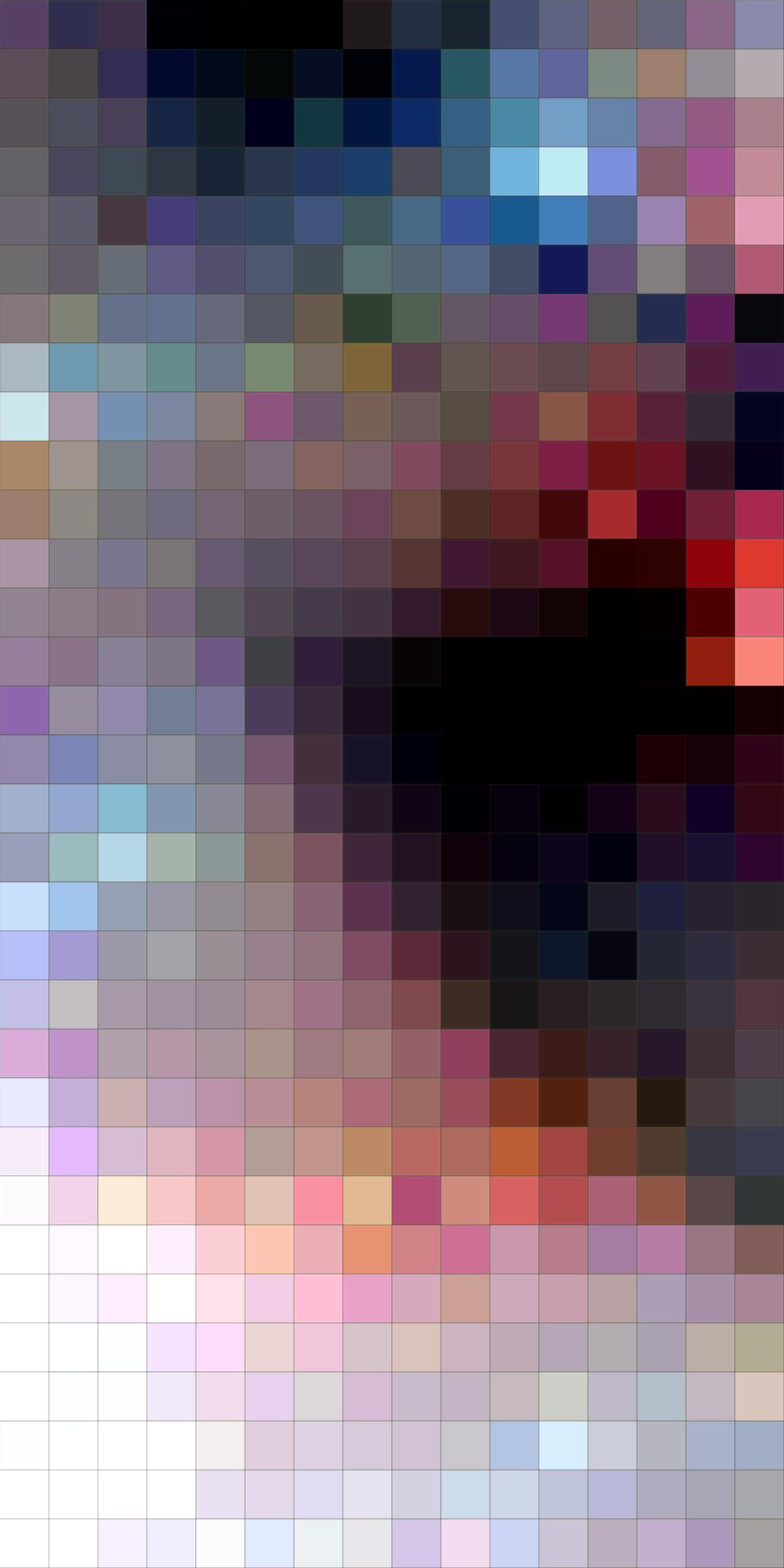}
	}
        \subfloat[Top=(27, 3)]{
		\includegraphics[width=.3\linewidth]{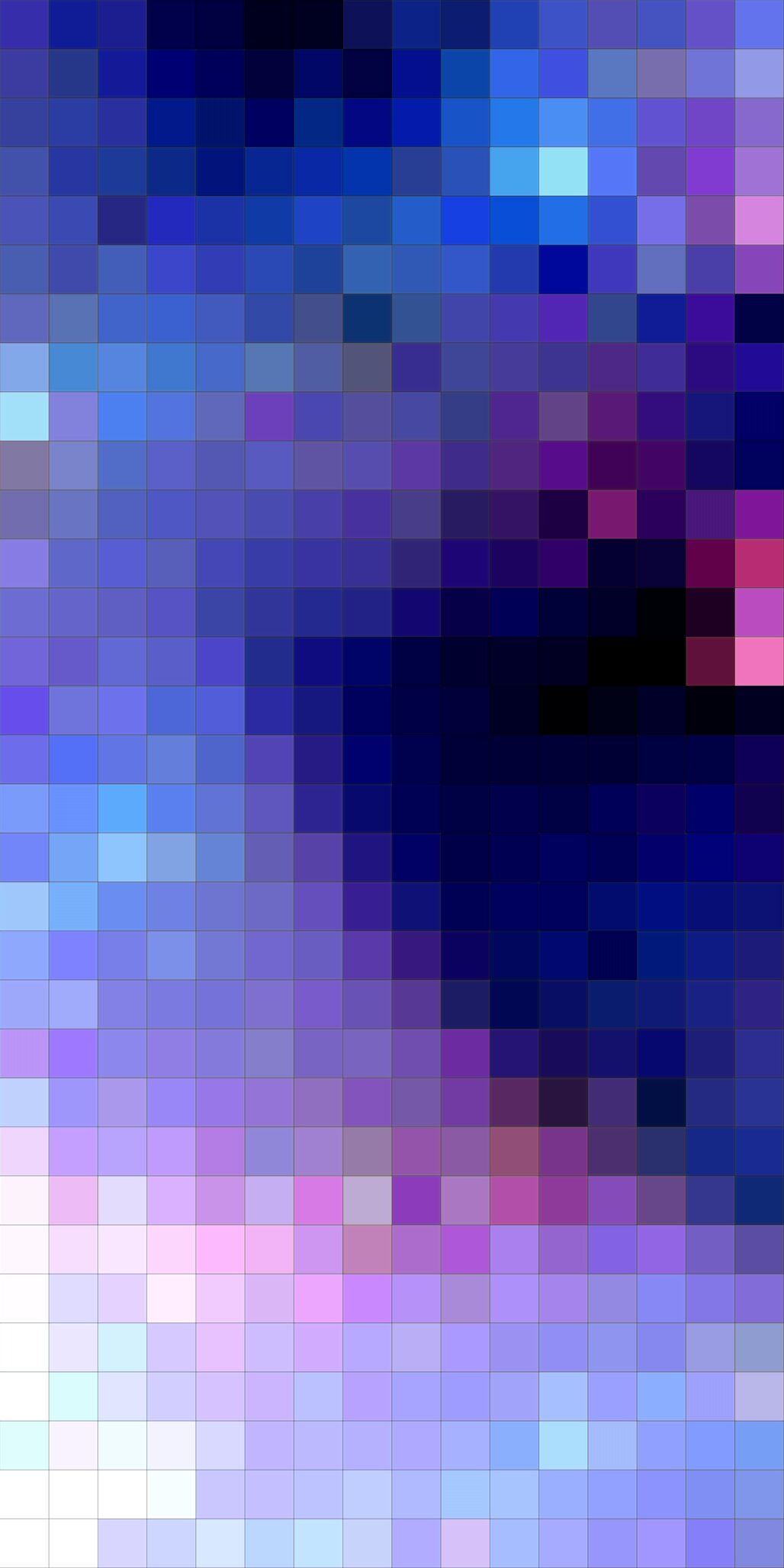}
	}
 \\
	\subfloat[Bottom=(0, 0)]{
		\includegraphics[width=.3\linewidth]{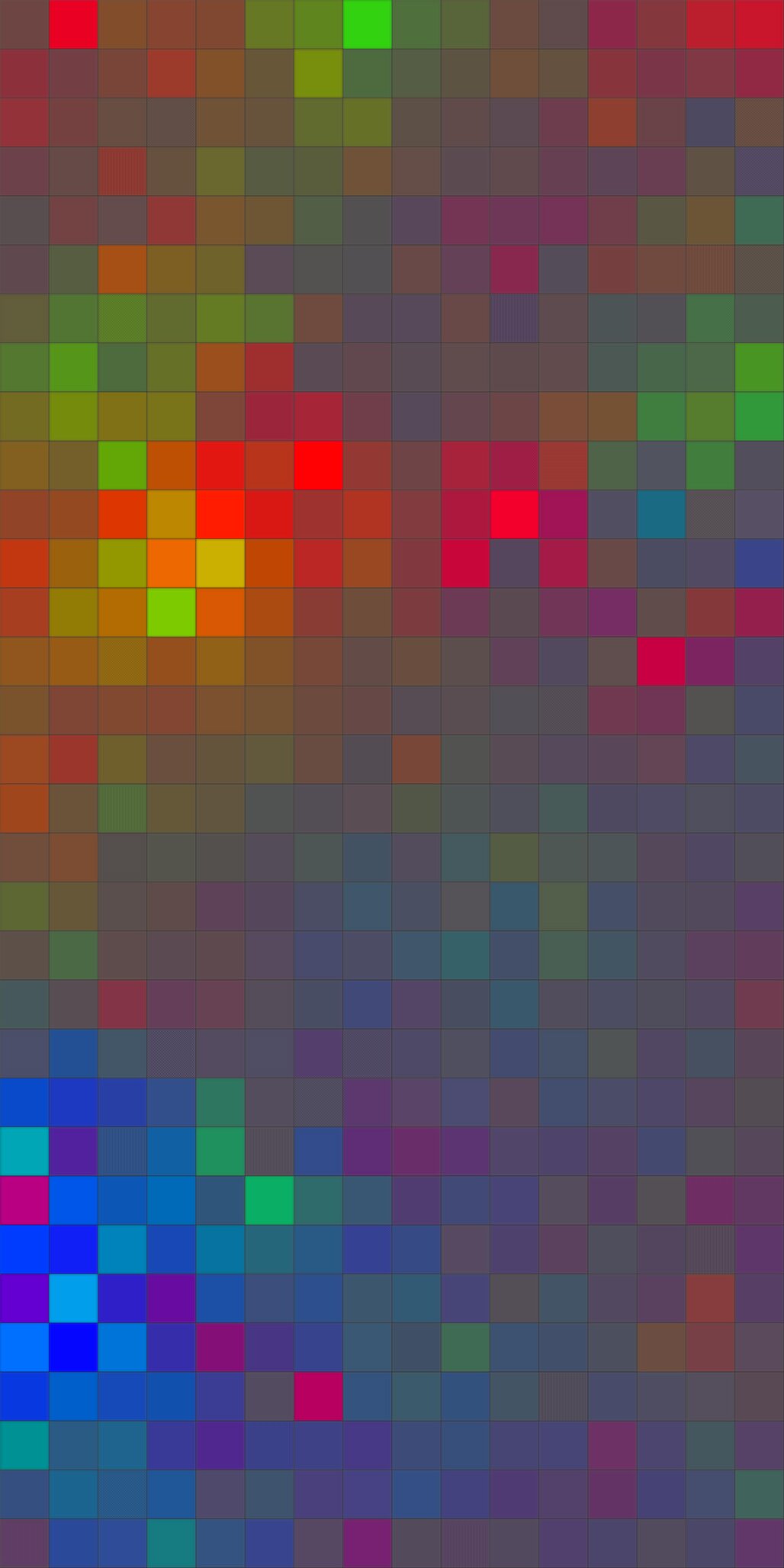}
	}
        \subfloat[Bottom=(18, 15)]{
		\includegraphics[width=.3\linewidth]{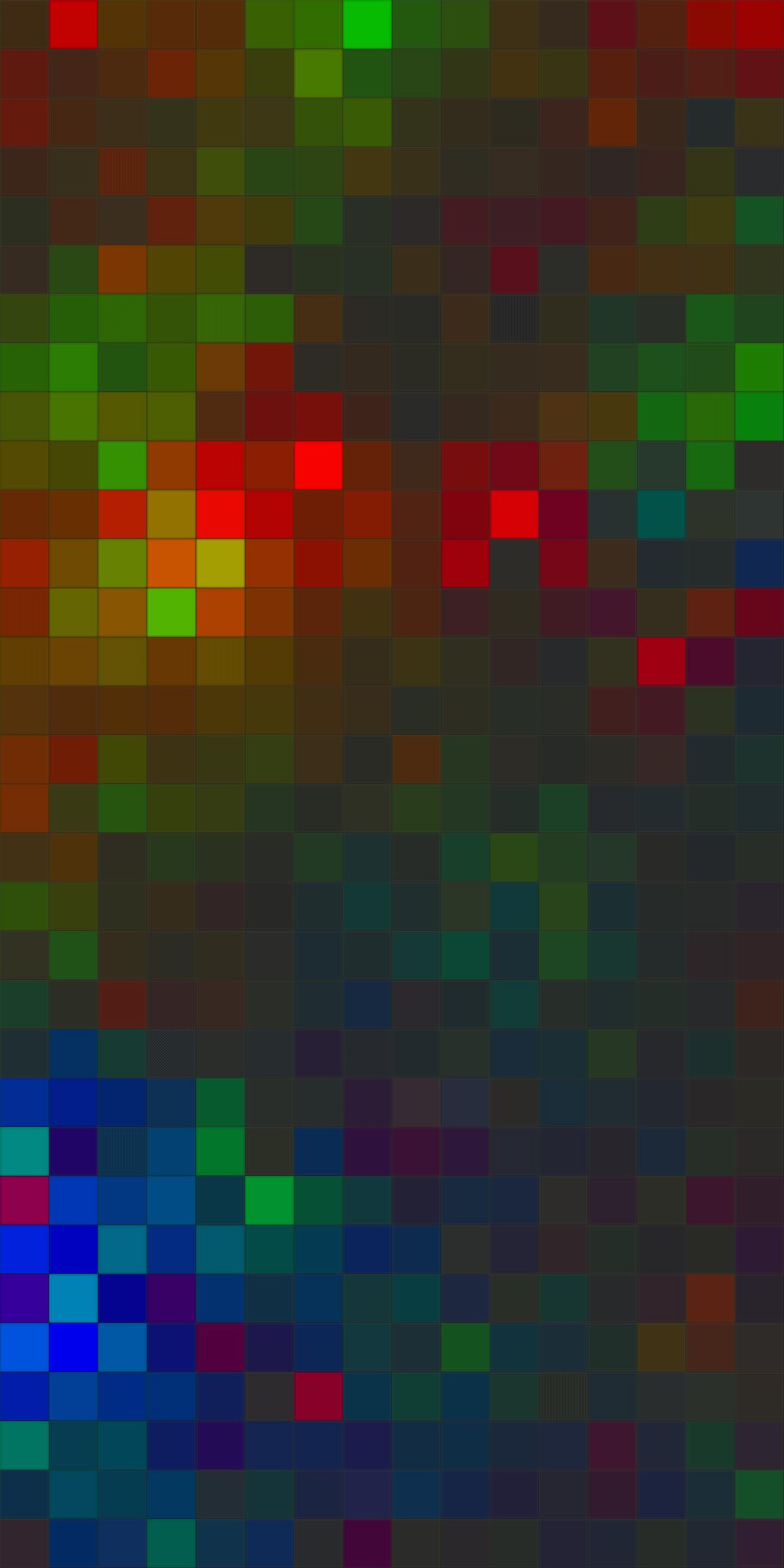}
	}
        \subfloat[Bottom=(27, 3)]{
		\includegraphics[width=.3\linewidth]{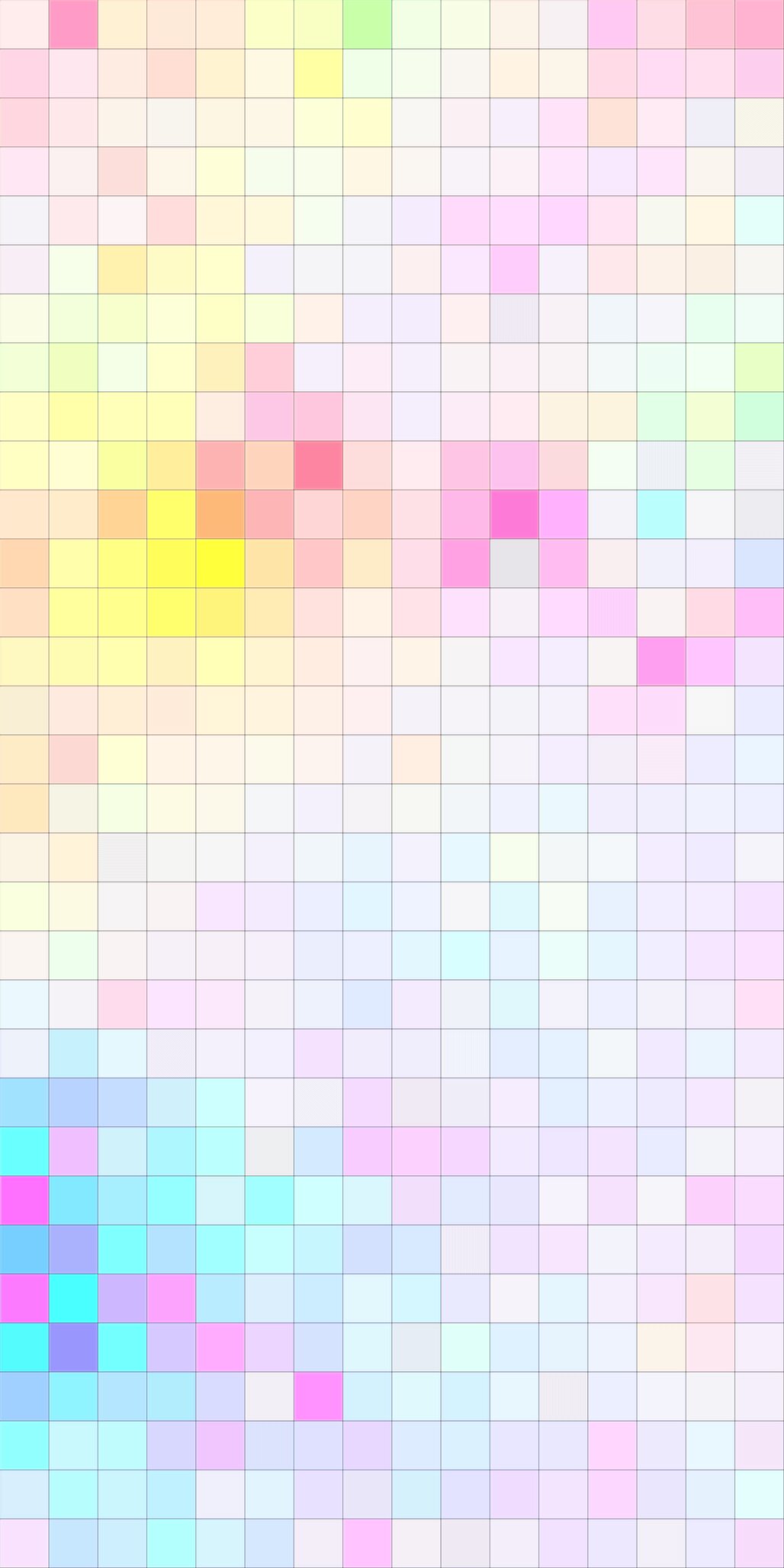}
	} 
	\caption{Codebook visualisations for VQ-VAE-2 trained on ImageNet with 2D-Hard KSOM. The description in the caption specifies which of the top or bottom latent representations is fixed to which coordinates.}
	\label{fig:vqvae2_codebook}
\end{center}
\end{figure}

\textbf{Grid Visualisation for VQ-VAE-2.}
In Figure \ref{fig:codebook_vis} in the main text, we visualise the 2D grid representation of the codebook for a VQ-VAE trained on CIFAR-10.
Here we show similar visualisations for VQ-VAE-2 trained on ImageNet.
One complication for such visualisations for VQ-VAE-2 is that it has two codebooks (``top'' and ``bottom'').
We therefore cannot exhaustively visualise all combinations.
We show a few of them by fixing all codes in either the top or bottom representations, and  varying the others.
Figure \ref{fig:vqvae2_codebook} shows the results.
While hue/value/saturation change for different values of fixed top or bottom code, we again observe many ``islands'' of similar colours grouped together.

\begin{figure}[t]
\begin{center}
	\subfloat[KSOM, shift top representation]{
  \hspace{5mm}
		\includegraphics[width=.95\linewidth]{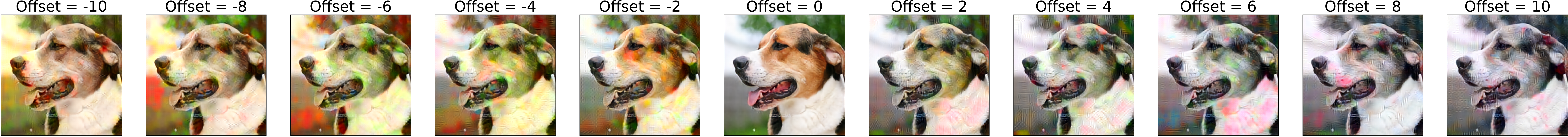}
	}
 \\
	\subfloat[KSOM, shift bottom representation]{
   \hspace{5mm}
		\includegraphics[width=.95\linewidth]{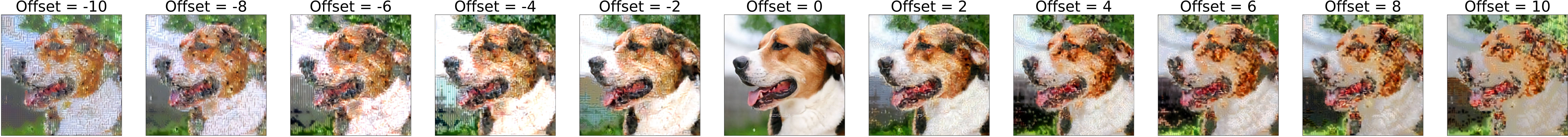}
	}
 \\
        \subfloat[EMA-VQ Baseline, shift top representation]{
   \hspace{5mm}
		\includegraphics[width=.95\linewidth]{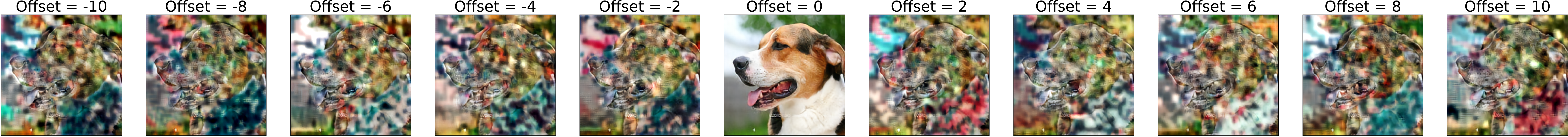}
	}
 \\
	\subfloat[EMA-VQ Baseline, shift bottom representation]{
   \hspace{5mm}
		\includegraphics[width=.95\linewidth]{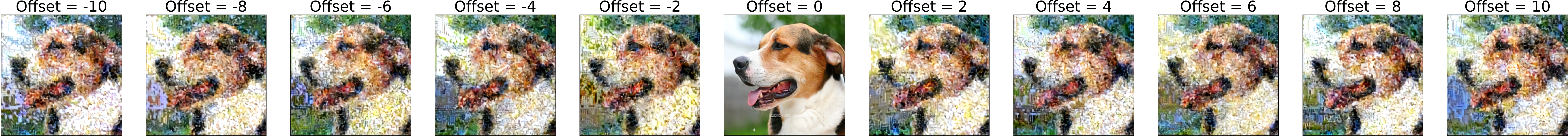}
	}
	\caption{Effects of perturbing only one of the top or bottom latent representations in VQ-VAE-2  (``dog'') }
	\label{fig:dog_top_bottom}
 \end{center}
\end{figure}

\begin{figure}[t]
\begin{center}
	\subfloat[KSOM, shift top representation]{
   \hspace{5mm}
		\includegraphics[width=.95\linewidth]{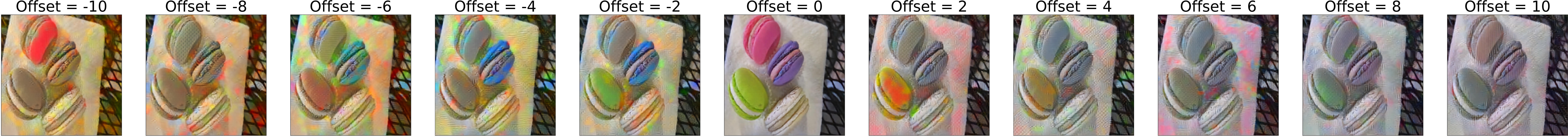}
	}
 \\
	\subfloat[KSOM, shift bottom representation]{
   \hspace{5mm}
		\includegraphics[width=.95\linewidth]{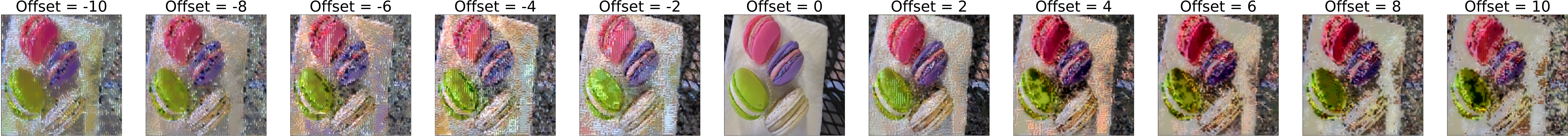}
	}
 \\
        \subfloat[EMA-VQ Baseline, shift top representation]{
   \hspace{5mm}
		\includegraphics[width=.95\linewidth]{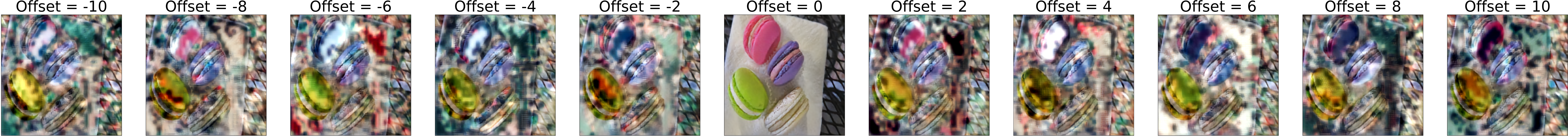}
	}
 \\
	\subfloat[EMA-VQ Baseline, shift bottom representation]{
   \hspace{5mm}
		\includegraphics[width=.95\linewidth]{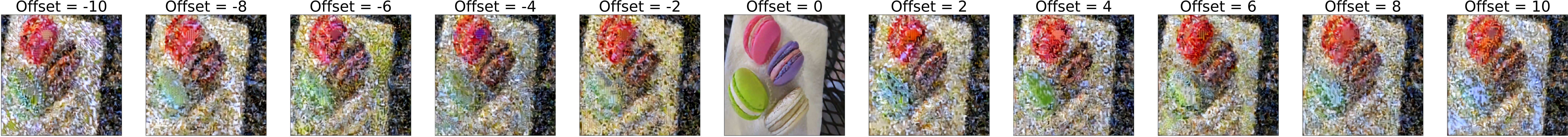}
	}
	\caption{Effects of perturbing only one of the top or bottom latent representations in VQ-VAE-2 (``macaron'')}
	\label{fig:macaron_top_bottom}
 \end{center}
\end{figure}

\textbf{Perturbing only one of the two latent representations in VQ-VAE-2.}
In Figure \ref{fig:shifting} in the main text, we apply an offset to all indices in the two latent representations (``top'' and ``bottom'') of VQ-VAE-2.
Here we show the effect of perturbing
only one of the two latent representations, i.e.,
we keep one of the ``top'' or ``bottom'' latent representations fixed (to that of a proper image), and apply offsets to the other one.
Figures \ref{fig:dog_top_bottom} and \ref{fig:macaron_top_bottom} show the corresponding results.
Since one of the two latent representations is fixed, the ``contents'' of the original image is somewhat preserved in all cases.
The comparison is interesting in the case where we perturb
the top latent code: in the KSOM case, the changes of top latent code seem to gradually affect the hue/value of the image.

\end{document}